\documentclass[journal]{IEEEtran}
\usepackage[utf8]{inputenc}
\usepackage{cite}
\ifCLASSINFOpdf
  \usepackage[pdftex]{graphicx}
  \graphicspath{{images/}}
\else
\fi
\usepackage[cmex10]{amsmath}
\usepackage{amssymb}

\usepackage{hyperref}

\usepackage[table,xcdraw]{xcolor}

\usepackage[normalem]{ulem}
\useunder{\uline}{\ul}{}
\usepackage{afterpage}
\usepackage{graphicx}

\ifCLASSOPTIONcompsoc
 \usepackage[caption=false,font=normalsize,labelfont=sf,textfont=sf]{subfig}
\else
 \usepackage[caption=false,font=footnotesize]{subfig}
\fi
\usepackage{booktabs}       % professional-quality tables
\usepackage{algorithm}
\usepackage[noend]{algpseudocode}
\usepackage{color}
\newcommand{\etal}{\textit{et al}.}

\newcommand{\norm}[1]{\left\lVert#1\right\rVert}

\newcommand*{\vertbar}{\rule[-1ex]{0.5pt}{2.5ex}}

\begin{document}

%\title{Real Time Monitoring of Large Scale Infrastructures using Depth Data}
\title{Understanding People Flow in Transportation Hubs}
\author{Jo\~ao~Carvalho,
        Manuel Marques, Jo\~ao P. Costeira
\thanks{Accepted in IEEE TRANSACTIONS ON INTELLIGENT TRANSPORTATION SYSTEMS, DOI:~\href{https://doi.org/10.1109/TITS.2017.2775285}{10.1109/TITS.2017.2775285}. 
    
    This work was supported by the Portuguese Foundation for Science and Technology, under the project FCT [UID/EEA/50009/2013]. Jo\~ao Carvalho is also supported by the FCT grant PD/BD/114429/2016.}% <-this % stops a space
\thanks{The authors are with the Institute for Systems and Robotics (ISR/IST), Instituto Superior T\'ecnico, Universidade de Lisboa, Portugal.}}% <-this % stops a space

\maketitle

\begin{abstract}
    In this paper, we aim to monitor the flow of people in large public infrastructures. 
        We propose an unsupervised methodology to cluster people flow patterns into the most typical and meaningful configurations. By processing 3D images from a network of depth cameras, we build a descriptor for the flow pattern. We define a data-irregularity measure that assesses how well each descriptor fits a data model. This allows us to rank flow patterns from highly distinctive (outliers) to very common ones. By discarding outliers, we obtain more reliable key configurations (classes).
    Synthetic experiments show that the proposed method is superior to standard clustering methods.
        We applied it in an operational scenario during 14 days in the X-ray screening area of an international airport. 
    Results show that our methodology is able to successfully summarize the representative patterns for such a long observation period, providing relevant information for airport management. 
    Beyond regular flows, our method identifies a set of rare events corresponding to uncommon activities (cleaning, special security and circulating staff). 
\end{abstract}

%------------------------------------------------------------------------- 
%\thispagestyle{fancy}

\section{Introduction}
\label{sec:intro}
In this article, we address the problem of crowd monitoring in public infrastructures. We propose a sensing and data processing framework which captures crowd occupancy in these large spaces, and identifies and classifies it into meaningful spatial patterns/classes. 
We tested it in a real-life scenario in the main security X-ray screening area at Lisbon's international airport (LIS), Portugal.

Airports are transportation hubs subject to strict service-level agreements (SLA), high security risks and high operational costs. These factors put great pressure on human and physical resources, calling for tight monitoring of passenger flow within the infrastructure.
In particular, the security and identification checkpoints are critical bottlenecks in the path between the check-in and the departure gate. Besides the risk of SLA violation, these bottlenecks have great impact on operational and commercial costs. 
Our work focuses on characterizing how passengers flow while waiting for inspection in these critical checkpoints. 

The sensing infrastructure we propose consists of a network of depth cameras that provide 3D data of the covered space. A set of pre-processing algorithms anonymously detect people in the 3D point cloud and compute the 3D space occupancy map.
\begin{figure}[t]
\centering
\includegraphics[width=0.48\textwidth]{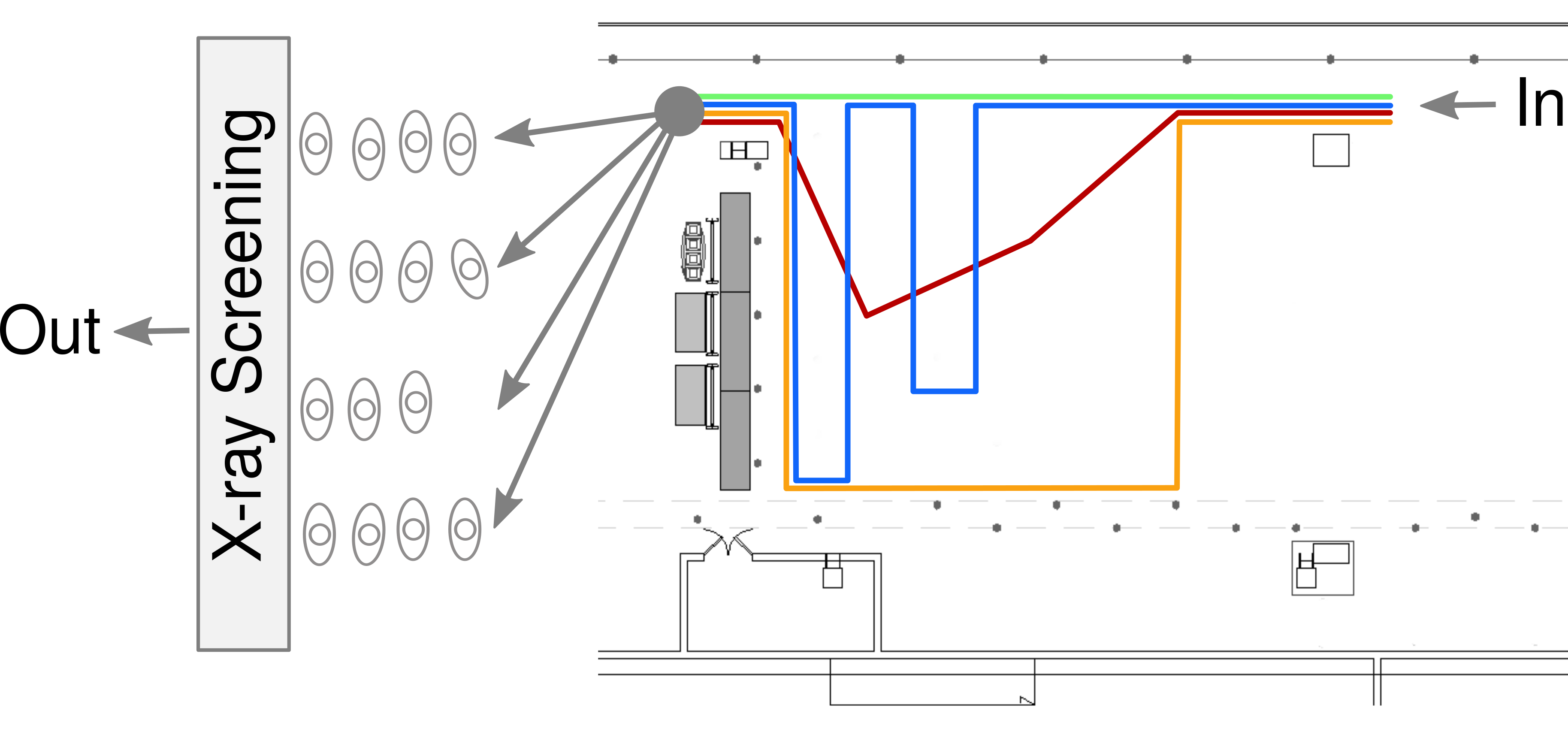}
\caption{X-ray screening queue area at the Lisbon international airport. The colored lines illustrate examples of paths in the queue area. The queue area is around $200\text{m}^2$.}
\label{fig:queue-modes}
\end{figure}
Fig. \ref{fig:queue-modes} displays the specific geometry of LIS X-ray area\footnote{The physical infrastructure described here no longer exists.}, where passengers enter from the right, wait for their turn and exit to the left towards individual X-ray booths. In such a large space---more than $200\text{m}^2$---several configurations emerge, shaped by inflow variations, processing capacity, planning protocols and particular events. 

Currently, the global status of the X-ray system is monitored using global counts of in-out passenger flow \cite{denman2015automatic,felkel2012comprehensive}. These global measures do not account for the internal state of the queue that reflects the instantaneous operating condition of the whole system.
Queue modeling (e.g., flow parameterization) is very difficult to do, despite some structure imposed by queue guides and a relatively controlled environment. X-ray queue configurations are set in an ad-hoc manner and often change according to discretionary decisions by operations personnel. 
Also, innumerable ``small'' local decisions impact the global system: metal detectors settings, protocols for suspicious luggage screening, passenger inspection or security personnel skills. 
All these aspects are hard to account for but affect the performance and overall state of the X-ray queue.

Our methodology for classification of passenger flow patterns identifies meaningful classes that encode space occupancy. Classes are defined by space occupancy patterns that appear regularly over time (\emph{regular} patterns) as well as sporadic patterns that correspond to ``atypical behaviours'' (\emph{irregular} patterns). 
The classes and its number are not known \emph{a priori} and the amount of outliers can be significative.
Because of the large amount of data, it is very difficult for the airport management to define the most relevant classes.

Building on previous work in the computer vision domain \cite{oat2016mfs}, we encode regularity/irregularity through linear subspace models that explicitly represent these regularity concepts \cite{soltanolkotabi2012geometric,elhamifar2013sparse}. Such subspaces are determined from data in an unsupervised way. 

Our methodology grants decision-makers with the much sought holistic measure of the ``operational state'' of the infrastructure. The methodology is general and applicable to a wide variety of public infrastructures such as railway stations or commercial shopping centers.

In summary, our main contributions are:
\begin{itemize}
    \item a 3D sensing infrastructure for queue data collection;
	\item a descriptor computed from 3D data of multiple cameras that captures queue patterns;
    \item an algorithm that explicitly models and identifies \emph{regular} and \emph{irregular} flow patterns and clusters the regular data into classes of ``typical'' operation modes.
\end{itemize}

By identifying regular ``operating states'' of the whole system, airport experts are able to rank and map them to  desirable performance indicators. 
Having these states, one can build individual and more accurate models for each of them using queueing theory, for example, seeking to improve management and security. The ultimate goal is to increase costumer satisfaction. 
This interaction with the management and operational staff is very important in order to understand the behavior of the whole system~\cite{chen2015survey}. Our methodology can be understood as a pre-processing stage whose information can feed the planning, scheduling and performance analysis systems of the airport.

\section{Previous Work}
\label{sec:related-work}

In complex environments such as an airport \cite{wu2013review}, the flow of passengers depends on a large number of variables: staff performance, passenger behaviour, hand luggage contents, sensitivity of metal detection systems, among others. 
Traditional approaches based on discrete event systems \cite{guizzi2009discrete,dorton2016effects} or queueing theory \cite{gilliam1979application,takakuwa2003modeling,de2013virtual} model airport activity as a function of several such variables. 
However, these methods are difficult to tune because they are highly dependent on variables that are unobservable or modeled with (unrealistic) stationary distributions (\emph{e.g.}, arrival and service rates). Also, these are based on data manually acquired within limited time frames or based on aggregated statistics.
Besides providing tools for real time data acquisition, our methodology identifies the stationary states of the queue, allowing the design of individual and more accurate models for each state.

To automatically acquire people flow data, several works track and count people with a single video camera \cite{albiol2009statistical,barandiaran2008real} or a single over-the-head RGB-D camera \cite{gao2016people,del2015versatile,fu2014scene}. 

In crowded scenes, instead of tracking people individually, several works propose solutions for activity detection or estimation of the number of people present in the scene.
For example, Convolutional Neural Networks (CNN) are employed in \cite{zhao2016crossing} to compute a per-pixel crowd counting map in order to estimate the number of people crossing a line.  
\cite{zhang2015cross} improves crowd counting by training a CNN using two related objectives, crowd density and count, to obtain a better local optimum for both.
Alternatively, \cite{zhang2016single} proposes a multi-column CNN to map an image to its crowd density map.
In \cite{ma2013crossing} and \cite{mukherjee2015unique}, the authors use features derived from optical flow to train a system for counting people in regions of interest at crowded places. 
The optical acceleration and histogram of optical flow gradients can be combined to detect abnormal objects or speed violation in pedestrian scenes \cite{nallaivarothayan2014mrf}.
A probabilistic approach is employed by Wang \etal, \cite{wang2009unsupervised}, to cluster moving pixels and video segments in order to model atomic activities and interactions in crowded places.
In \cite{roshtkhari2013online}, the detection of abnormal behaviors is done with a spatio-temporal analysis of a densely sampled video volumes.

Clustering individual trajectories is another common approach to analyze the flow of people.
In \cite{cheriyadat2008detecting}, feature point tracks are clustered and dominant trajectories are identified by fitting polynomials to cluster mean points.
Approach \cite{lei2016robust} proposes the Robust K-means algorithm for clustering data that is less sensitive to the initialization of the $K$ clusters.
K-means and agglomerative clustering are used in \cite{kalayeh2015understanding} to cluster trajectories into common patterns. 
Alternatively, in \cite{morris2008learning}, trajectories are clustered with Fuzzy C-means and modeled with HMM, for trajectory analysis. Also, local distance and similarity measures are frequently employed to cluster and analyze trajectories and flow data \cite{liao2005clustering, hou2016repeatability}.

In \cite{wu2011real}, the authors propose a system for monitoring queues by tracking people using a network of over-the-head video cameras. 
A solution to integrate existing monitoring technology is proposed in \cite{denman2015automatic}. It studies methodologies for people counting and individual tracking for queue monitoring and behavior analysis. 

In our scenario, the placement of the sensors is limited by the infrastructure, leading to large scene perspective distortion, people occlusion and rendering the registration of video cameras unfeasible. 
Moreover, due to privacy concerns, the use of video cameras is not allowed in many locations, including the security checkpoint at Lisbon Airport.
This precludes the use of the referenced approaches, which, besides using RGB cameras, assume: high ceilings, proposing over-the-head solutions \cite{gao2016people, del2015versatile,wu2011real}; access to airport sensing infrastructure \cite{denman2015automatic}; existence of reliable individual trajectories \cite{cheriyadat2008detecting,kalayeh2015understanding,morris2008learning, candamo2010understanding}; absence of outliers% in the data
, as irregular patterns \cite{lei2016robust}; and large amounts of training data \cite{zhang2015cross, zhang2016single, zhao2016crossing}.

To the best of our knowledge, there is no work proposed for unsupervised identification and classification of people flow patterns from depth data.
Our approach is applicable in a large range of scenarios because it does not require a tracking scheme.
Also, we bring the novelty of testing our methodology with $14$ days of real data acquired at an international airport.

\section{Sensing Infrastructure and Data Descriptor}
\label{sec:sensing}
To characterize the X-ray queue area, we want to capture the occupancy pattern of the passengers over time. 
As we will see in Section \ref{sec:params}, counting people at the entrance or exit of the queue is not enough to understand the state and shape of the queue. Also, individually tracking people in such a large area is unreliable, due to frequent occlusion of passengers. We propose a descriptor 
that captures the dynamics of the space and is robust to misdetections.
We present here a sensing infrastructure and, based on 3D data processing, the proposed descriptor.

\begin{figure}[tbh]
\centering
\includegraphics[width=0.38\textwidth]{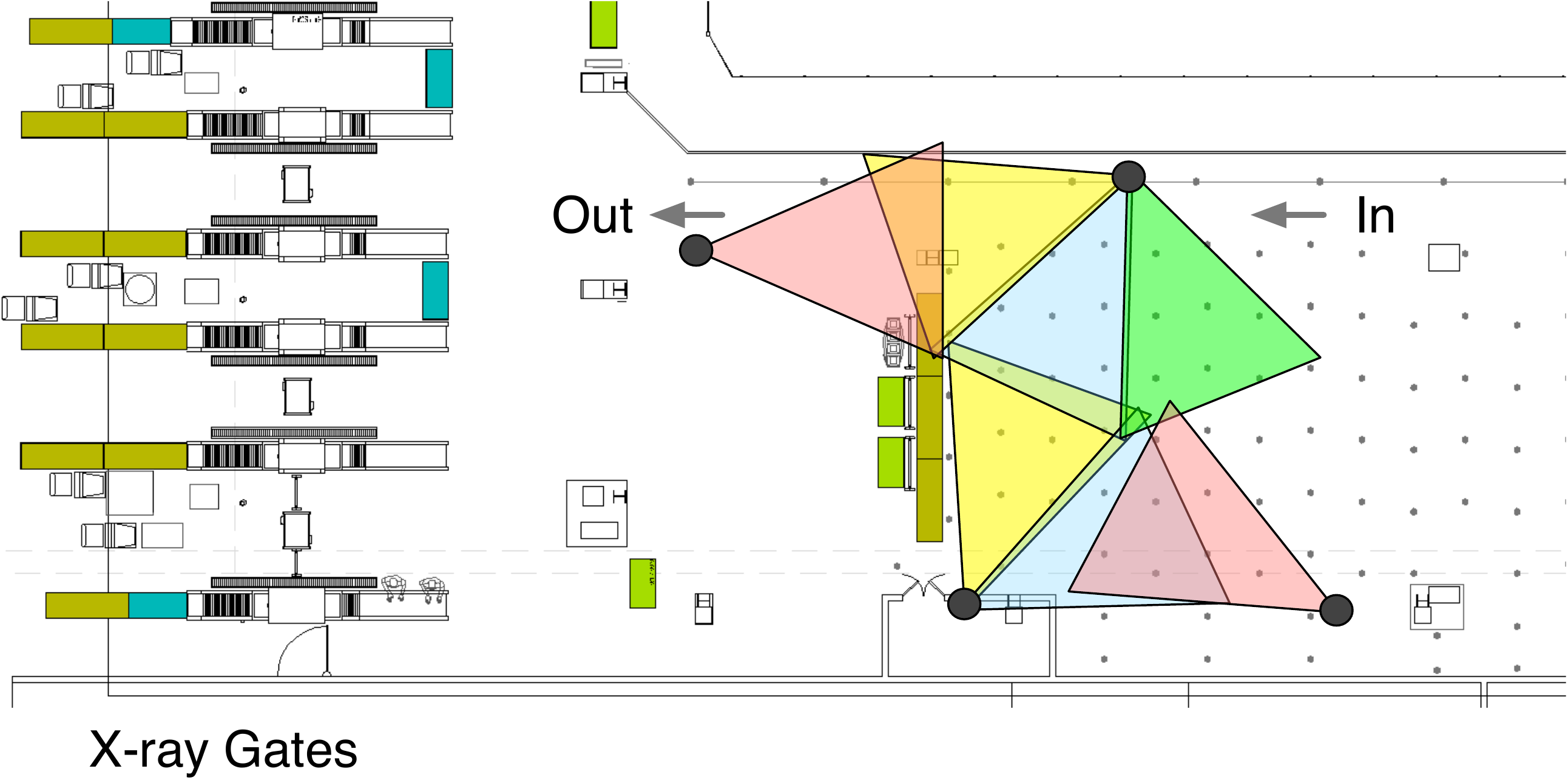}
\caption{Camera setup in the X-ray queue with corresponding field of view (FOV). Each colored triangle represents the FOV of a different camera. Passengers enter from the right and exit on the left.}
\label{fig:rx-cams-fov}
\end{figure}

\begin{figure}[tbh]
\centering
\subfloat[Depth image]{\includegraphics[width=0.14\textwidth]{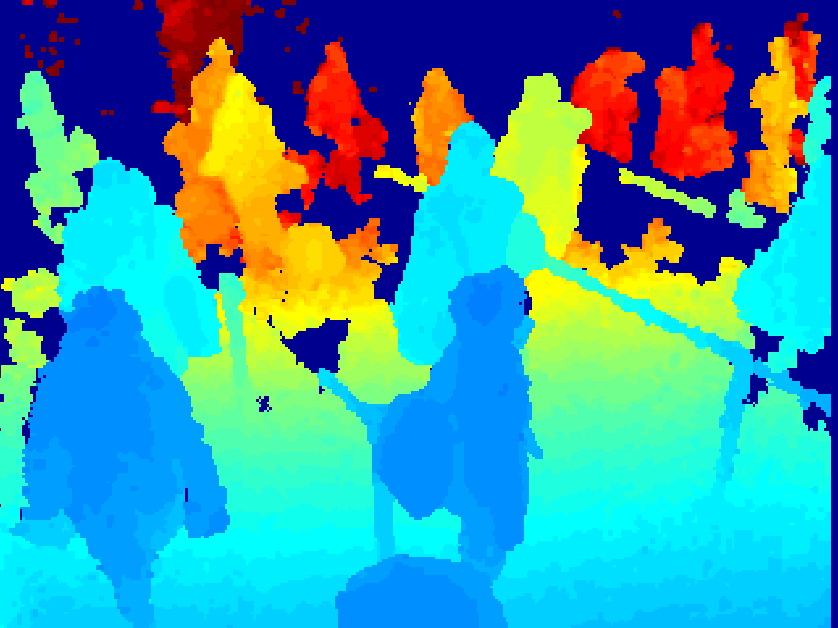}
\label{fig:img-depth}}
\subfloat[3D point cloud of people standing]{\includegraphics[width=0.3\textwidth]{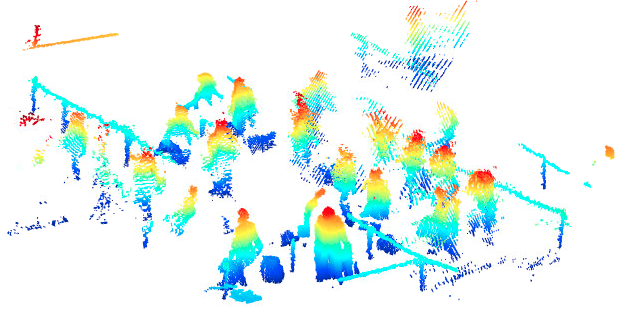}
\label{fig:pcl-rx-pre}}
\caption{Depth image and 3D representation: (a) Depth image from one of the cameras placed at the X-ray queue area at the Lisbon International Airport. The value of each pixel represents the depth of the object rather than its brightness or color. The colors are indexed to the distance to the camera, from blue (closer) to dark red (further). Dark blue represents areas not visible by the camera. (b) 3D representation of the space. Colors range from dark blue, for points closer to the ground plane, to dark red, for points further from the ground plane.}
\label{fig:3d-data}
\end{figure}

Fig. \ref{fig:rx-cams-fov} depicts the network of depth cameras installed to cover the final stage of the queue area.
Cameras are calibrated and 3D point clouds are registered into a global reference frame. 
So, from depth images provided by the seven cameras (Fig. \ref{fig:img-depth}), we build a global 3D representation of the space (Fig. \ref{fig:pcl-rx-pre}).
Since we have a noisy sensor and multiple objects in the foreground, we used a procedure we developed in \cite{carvalho2016detecting} to detect people in 3D point clouds (Section \ref{sec:detection}).
Using the results of this detection procedure, we compute a ground occupancy map and use it as a descriptor for the passenger flow (Section \ref{sec:occ_map}).
\subsection{People Detection}
\label{sec:detection}
\begin{figure}[htb]
\centering
\subfloat[People detections]{\includegraphics[width=0.27\textwidth]{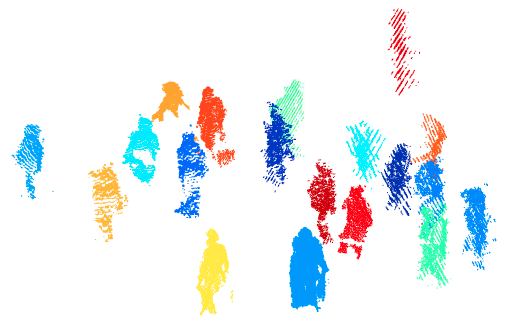}
\label{fig:pcl-rx-detections}}
\subfloat[People centroids project on the floor]{\includegraphics[width=0.15\textwidth]{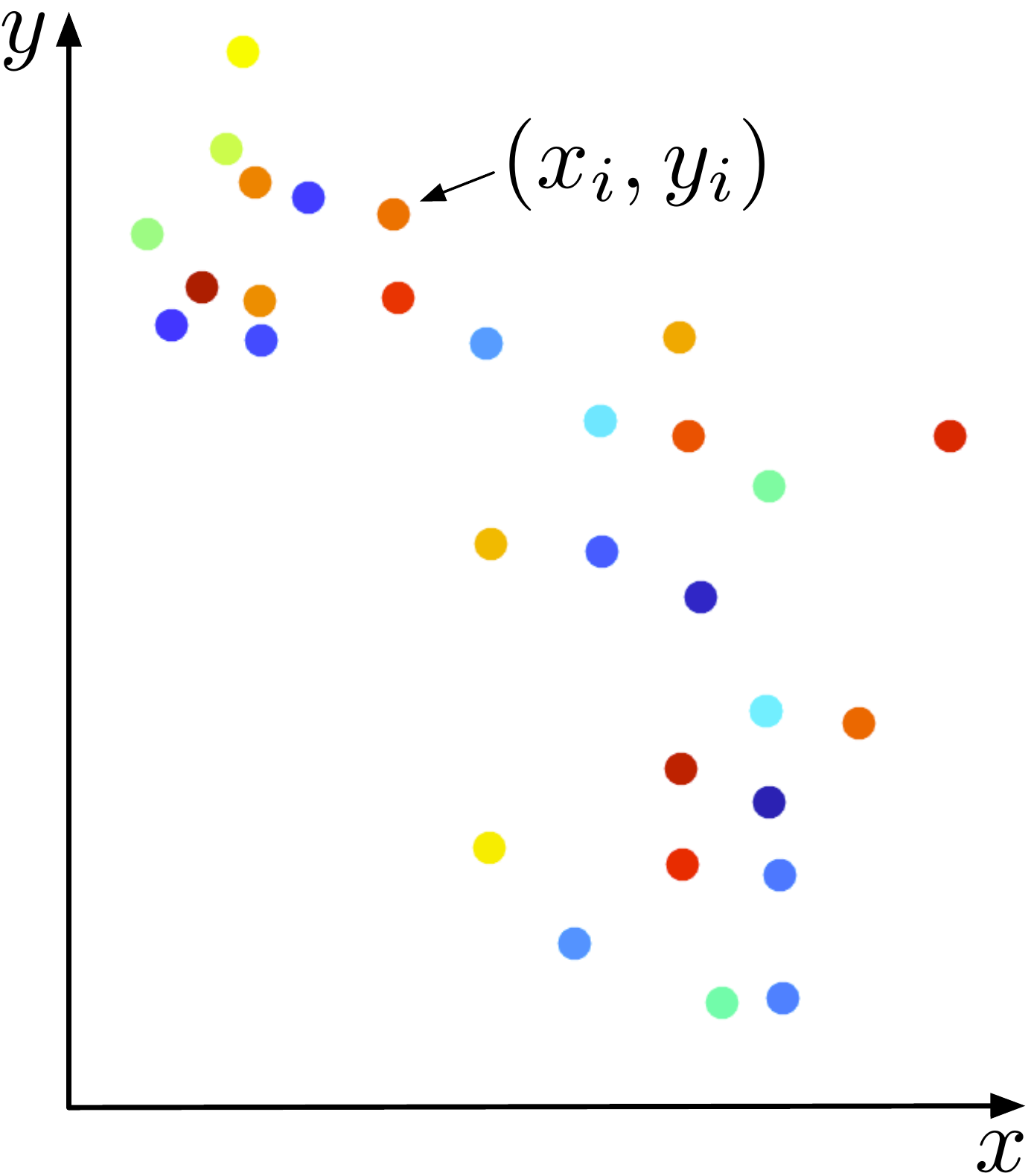}
\label{fig:people-centroids}}
\caption{People detection method results: (a) point cloud of the people detection, with one color per person; (b) centroids $(x,y)$ of each detection.}
\label{fig:detection-out}
\end{figure}

The method takes a 3D point cloud as input (Fig. \ref{fig:pcl-rx-pre}) and outputs the 3D points and centroids corresponding to each person. Fig. \ref{fig:pcl-rx-detections} shows the 3D point clouds classified as ``person'', with one color per person, and Fig. \ref{fig:people-centroids} shows the centroid of each person projected on the ground plane.

The 3D space is scanned with a fixed size box and the point cloud falling within that volume is classified as \emph{person} or \emph{not person}.
The classifier, based on random trees, was trained with point clouds including people in different poses relative to the camera.
From each candidate point cloud, we compute its height, volume and the area occupied by the projection into the ground plane.
The output of this method is the labeling of the point cloud and the corresponding centroids in ground coordinates, $(x_i, y_i)$ for person $i$, as shown in Fig. \ref{fig:detection-out}.
For further details, including training process and performance, see \cite{carvalho2016detecting}.
\subsection{Queue Pattern Descriptor}
\label{sec:occ_map}
\begin{figure}[hbt]
\centering
\subfloat[]{\includegraphics[width=0.14\textwidth]{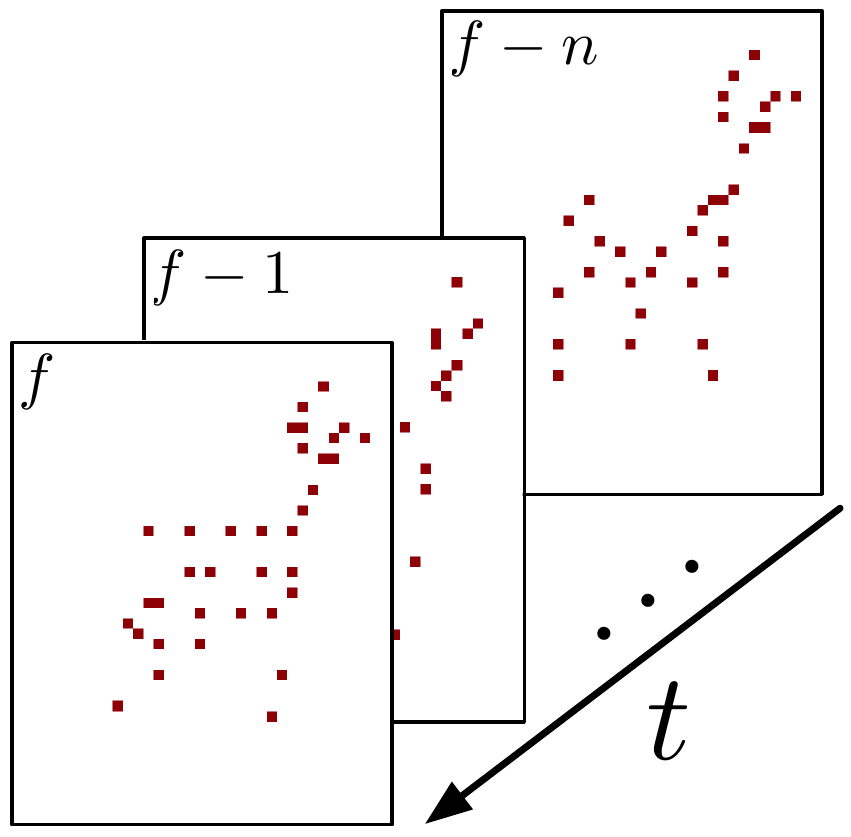}
\label{fig:binary-buffer}}
\hspace{3mm}
\subfloat[]{\includegraphics[width=0.1\textwidth]{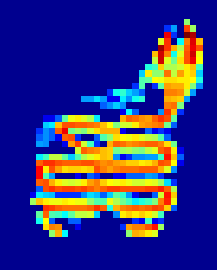}
\label{fig:occ-map}}
\caption{Occupancy maps: (a) buffer of binary maps with dark red cells if occupied and white otherwise; (b) occupancy map for $5min$ sized buffer. The colors range from dark blue, for the null value, to dark red, for the maximum value of the map.}
\end{figure}

We characterize the shape of the pedestrian flow using an occupancy map as data descriptor for the queue state. 
We build this occupancy map from the centroids $(x_i, y_i)$ output from the people detection method.
The floor is discretized into a fixed size grid and we obtain a binary map where each cell of the map is set to $1$ if occupied and $0$ otherwise. 
A cell is occupied if a centroid falls into it.
Fig. \ref{fig:binary-buffer} shows a set of these binary maps, from frame $f-n$ to $f$, where each dot corresponds to a occupied cell in the map.

The occupancy map of a given time period is the spatio-temporal average of the set of binary maps for that same period\footnote{A map can be instantaneous or integrated into longer periods, such as $30s, 1min, 5min, 1h.$}. 
We could apply other filters to the buffer of binary maps, however, we found that using the spatio-temporal average successfully captures the shape of pedestrian flow.
Fig. \ref{fig:occ-map} shows an example of an occupancy map. 
This map is a vector in $\mathbb{R}^{d}$ that evolves in time, depending on the space usage, where $d$ is the number of spatial cells of the map. 
\section{Queue Pattern versus Throughput}
\label{sec:params}
In this section, we show that using passenger counts \cite{denman2015automatic}, \cite{felkel2012comprehensive}, is not enough to characterize the occupancy of the area being monitored.

\begin{figure}[bht]
\centering
\includegraphics[width=0.23\textwidth]{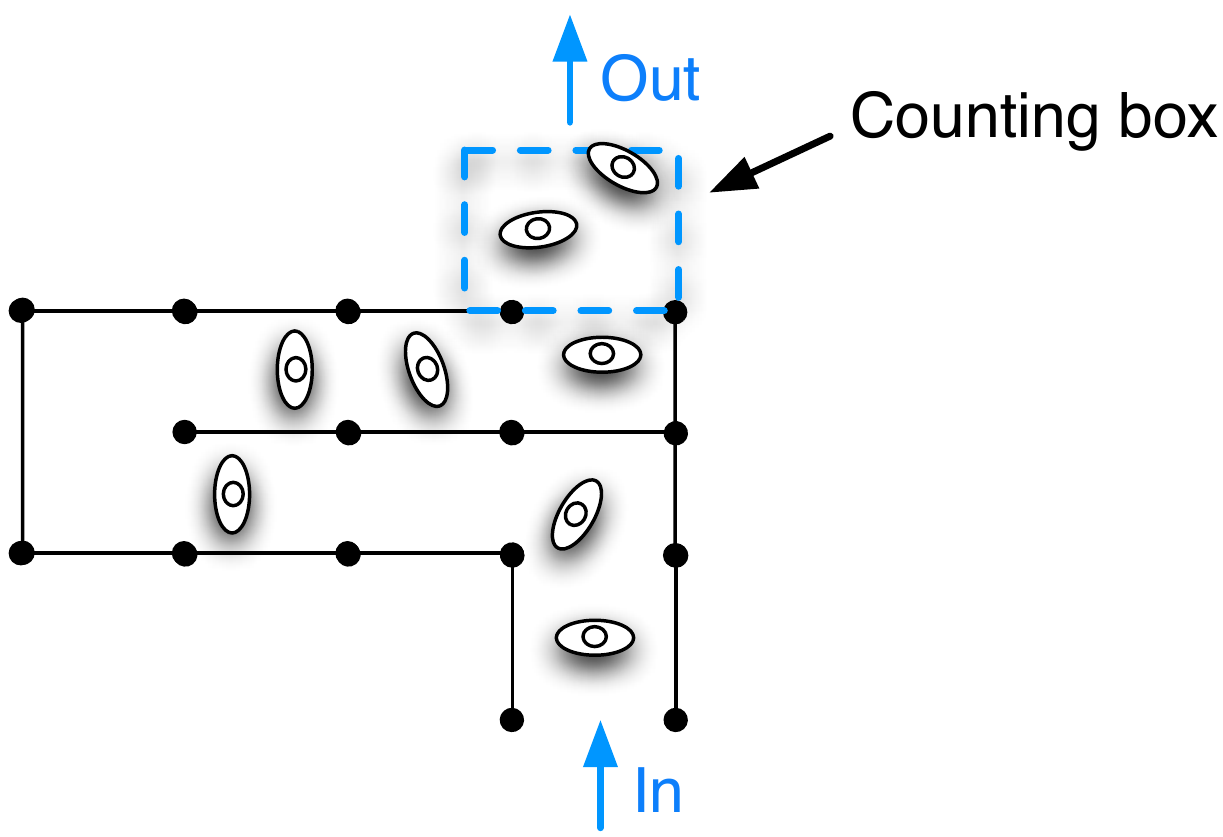}
\caption{Illustration of a counting box at the exit of a queue.}
\label{fig:queue-box}
\end{figure}

\begin{figure}[bth]
\centering
\subfloat[]{\includegraphics[width=0.34\textwidth]{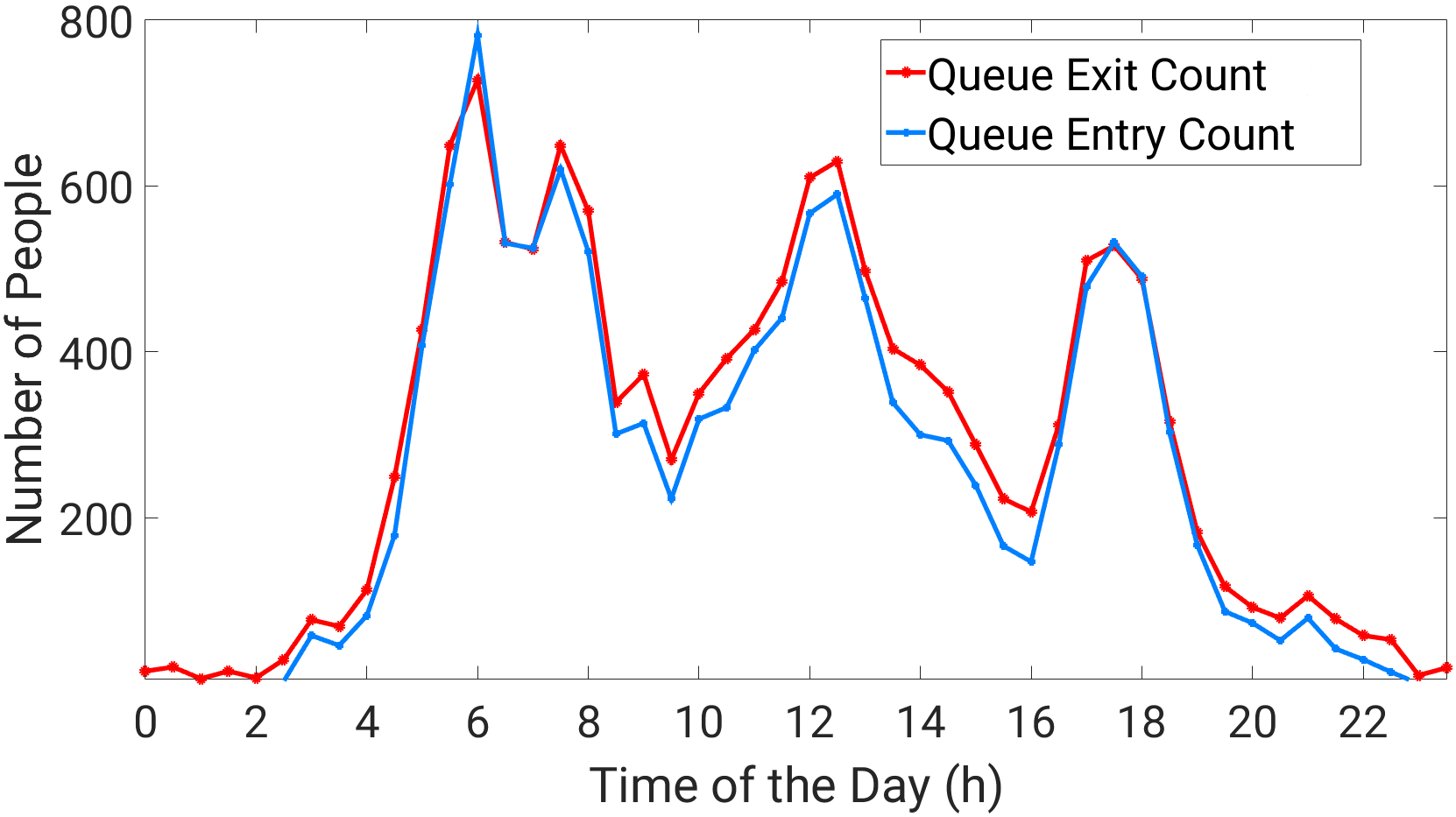}
}
\caption{Number of people passing by the X-ray queue detected with our approach versus ticket validation count. Differences between the two are due to the ground truth being based in boarding pass validation at the entrance of the queue and the count being relative to all the people, including staff, at the exit of the queue. Our detection methodology is able to capture the outflow variation.}
\label{fig:dashboard_valid}
\end{figure}

Based on the people detection procedure, we are able to count people passing in a small area at the exit of the queue (Fig. \ref{fig:queue-box}). 
In this scenario, we have $4\%$ error, similar to the classification error obtained in the original paper \cite{carvalho2016detecting}.
Fig. \ref{fig:dashboard_valid} shows the number of passengers at the entrance (blue line) and at the exit (red line) of the queue during one day. 
The passenger count at the entrance is provided by the boarding pass scanning system. However, at the queue exit, our system counts people without distinction between passengers and staff, therefore, the difference between the two counts is due not only to the detection error, but also due to staff passing the exit and the delay between the two counting systems (one is at the entrance and the other at the exit).
Despite these sources of error, we can see in Fig. \ref{fig:dashboard_valid} that our detection procedure successfully captures the outflow of the queue.
However, the count of passengers is not sufficient to identify the queue state.

\begin{figure}[bth]
\centering
\subfloat[]{\includegraphics[width=0.38\textwidth]{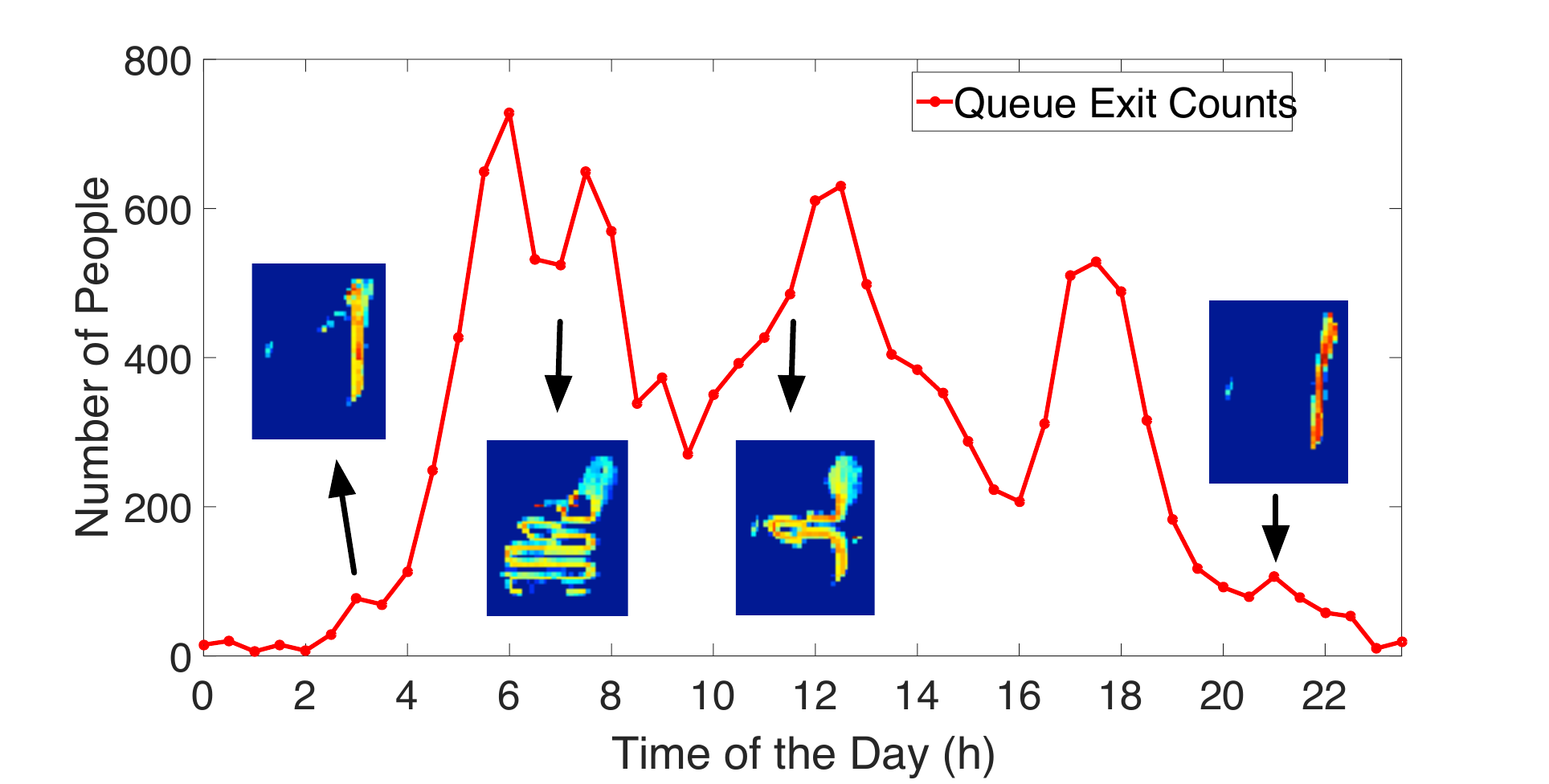}
}\\
\caption{X-ray exit count versus occupancy maps. Similar counts may correspond to different queue configurations.}
\label{fig:countings-maps}
\end{figure}

Fig. \ref{fig:countings-maps} shows four occupancy maps corresponding to four time instants. 
The periods around 07:00$am$ and 12:00$pm$ have very similar counts, however, the queue configurations are different. 
The state of the queue is dependent on many factors, including, for example, the relation between the queue inflow, service performance and the number of X-ray operating gates. 
Since these indicators are not available, the outflow of the queue is not enough to determine the queue/service state. Our solution relies on the occupancy map to capture the usage of the space.
\section{Detecting Queue Configurations}
\label{sec:charact}
In the previous sections, we described our sensing setup composed by a network of depth cameras that provides 3D data. 
With this data, we are able to detect people and compute an occupancy grid map -- a descriptor that does not rely on tracking to encode people movements.

\begin{figure}[tbh]
\centering
\includegraphics[width=0.43\textwidth]{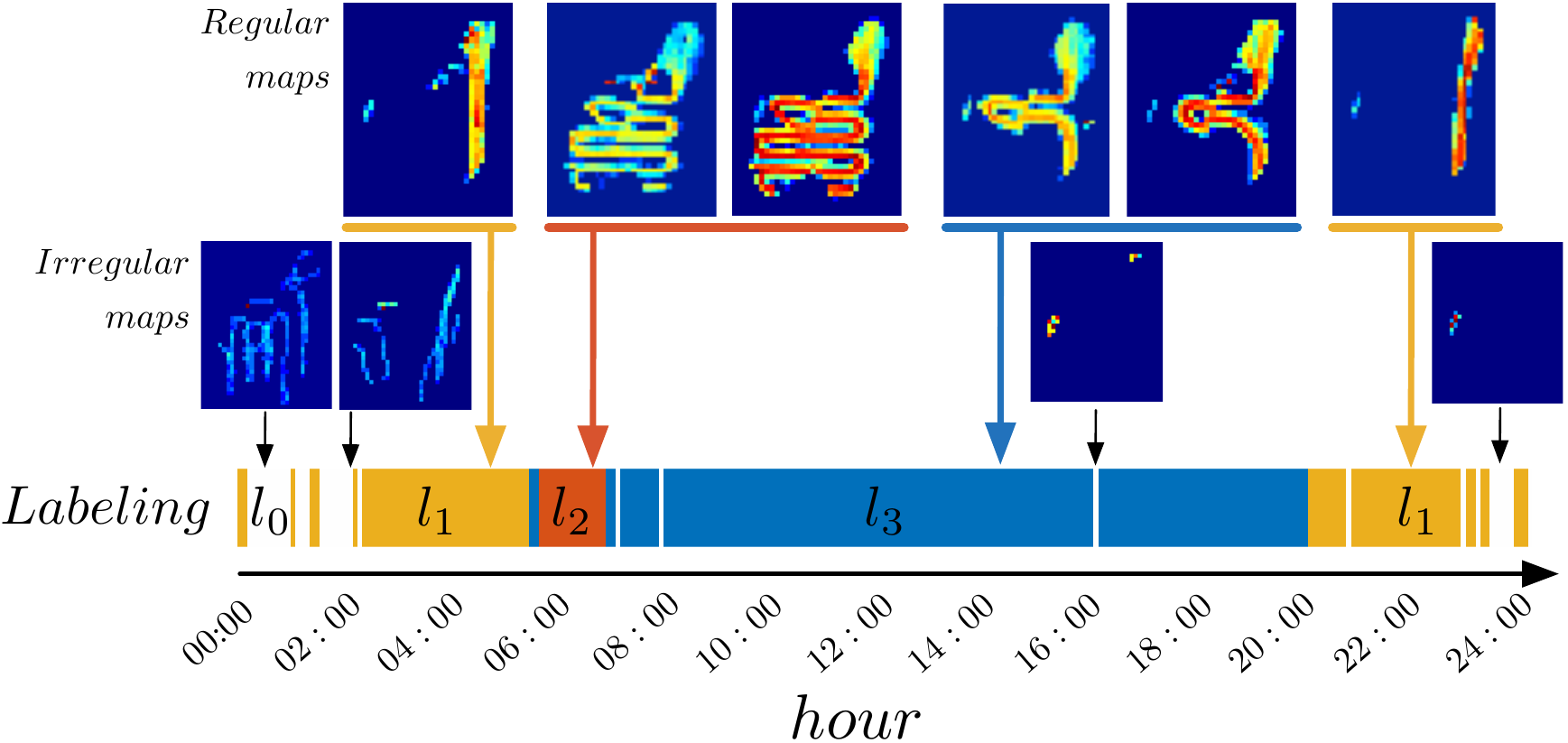}
\caption{This figure illustrates the goal of out methodology: going from occupancy maps to labels corresponding to the different queue configurations. Here, we illustrate three regular clusters ($l_1$ to $l_3$) and a cluster with irregular patterns ($l_0$, white label). Each cluster is composed by several similar patterns.}
\label{fig:goal}
\end{figure}

With the occupancy maps for a given period of time, we aim to cluster them into representative classes, where each class corresponds to a different queue configuration. 
Fig. \ref{fig:goal} illustrates this goal, with similar patterns grouped in the same cluster. 
Each cluster is represented in different color and with label $l_i$, where $i$ is the class index.
In this example, we illustrate three clusters with regular patterns ($l_1$ to $l_3$) and a white label ($l_0$) for irregular patterns.
Since we do not know the classes \emph{a priori}, a supervised approach is not adequate. 
On the other hand, the traditional unsupervised clustering methods do not guarantee a solution to our problem, as they require the number of classes to be known \emph{a priori}, and do not deal with outliers. 
As Fig. \ref{fig:goal} shows, we may have a considerable percentage of outliers in the data (white label, $l_0$). 

We model a typical pattern as a linear combination of other maps in the same class and compute the combination coefficients by solving a convex optimization problem. 
Because typical patterns appear regularly over time, we draw inspiration from the \emph{self-expressiveness} property~\cite{soltanolkotabi2012geometric,elhamifar2013sparse}, which states that a point in a subspace can be represented as a linear combination of points in the same subspace.
So, each class is modeled as a subspace, where each pattern of a given class is a point in the corresponding subspace. 
Solving this convex optimization problem, we have a measure of irregularity that assesses how distant a pattern is from the linear model. If this measurement is too high, the pattern is considered irregular (outlier).

\begin{figure}[tbh]
\centering
\includegraphics[width=0.4\textwidth]{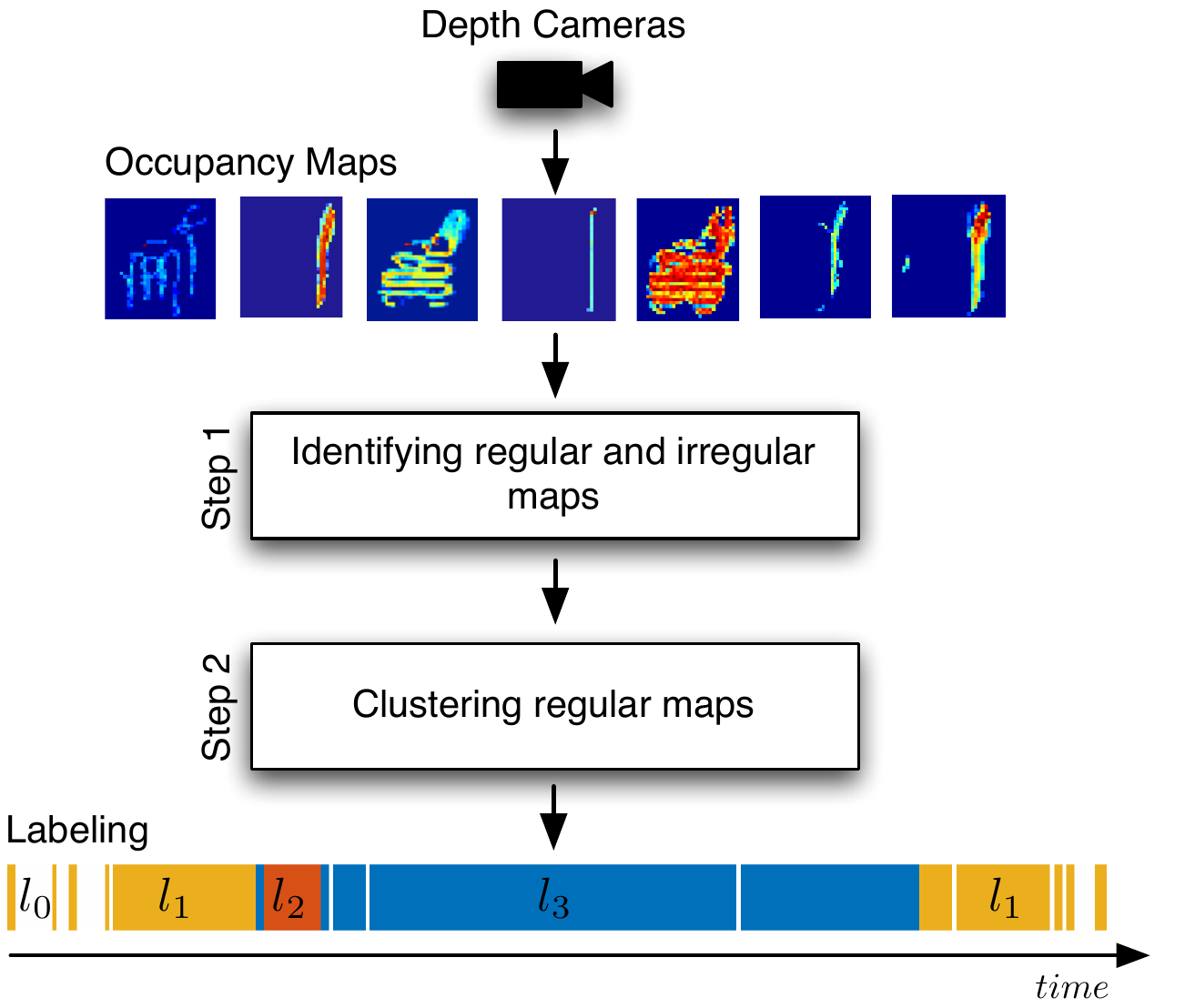}
\caption{Diagram summarizing the two main steps of the proposed queue classification methodology.}
\label{fig:method}
\end{figure}

Our method consists of two main steps: \emph{identifying regular and irregular maps}; and \emph{clustering regular maps}. 
Fig. \ref{fig:method} summarizes the methodology.
\subsection{Identifying Regular and Irregular Maps}
\label{sec:mfs}
Consider the data matrix
\begin{equation}
    \mathbf{X}=
\begin{bmatrix}
\vertbar & \vertbar & & \vertbar\\
\mathbf{x}_1 & \mathbf{x}_2 &  \dots  & \mathbf{x}_m \\
\vertbar & \vertbar &  & \vertbar \\
\end{bmatrix}
\in \mathbb{R}^{d\times m},
\end{equation}
where column $\mathbf{x}_i$ is a vectorized occupancy map, $d$ is the data dimension and $m$ is the number of maps\footnote{Bold capital letters, $\mathbf{A}$, represent matrices. Bold lower-case letters, $\mathbf{a}$, represent column vectors. Bold lower-case letters with subscript, $\mathbf{a}_i$, represent the $i^{th}$ column of matrix $\mathbf{A}$. Scalars are denoted by non-bold lower-case letters, $a$. The scalar element in row $j$ and column $i$ of matrix $\mathbf{A}$ is denoted by a non-bold lower-case letter with two subscripts, $a_{ji}$.}.

A map $\mathbf{x}_i$ is represented by a convex combination of other maps, being modeled as
\begin{align}
    &\mathbf{x}_i = \sum_{j=1, j\neq i}^m \mathbf{x}_j c_{ji} \label{eq:map-model}\\
    &\sum_{j=1, j \neq i}^{m} c_{ji} = 1, \quad c_{ji} \geq 0 \nonumber 
\end{align}
where $c_{ji}$ is the coefficient that relates map $j$ with map $i$.
The convex combination ensures that the map $\mathbf{x}_i$ is in the convex hull of other maps and, because of $c_{ji} \geq 0$, that the maps in the convex hull have similarities with $\mathbf{x}_i$.
Because a map should be expressed as a combination of maps belonging to the same cluster/subpace, our goal is to find the subspaces that best explain all the data.
In matrix form, the model is given by
\begin{equation}
    \mathbf{X}=\mathbf{XC} \label{eq:model-matrix}
\end{equation}
with
\begin{equation}
   \mathbf{1}_m^T \mathbf{C} = \mathbf{1}_m^T\mathrm{,}\hspace*{3mm} \mathbf{C} \geq \mathbf{0}_{m\times m}\mathrm{,}\hspace*{3mm} diag(\mathbf{C}) = 0 \nonumber
\end{equation}
where $\mathbf{C} \in \mathbb{R}^{m\times m}$ is the coefficients matrix and $\mathbf{1}_m$ is a vector of ones with dimension $m$. 
Imposing the null diagonal of $\mathbf{C}$, $diag(\mathbf{C})=0$, excludes the trivial solution of a map explaining itself.
The irregularity of a map is measured by how much the reconstruction of that map violates the linear model.
This is given by 
\begin{equation}
    \mathbf{\mathcal{I}}_i=\norm{\mathbf{x}_i-\mathbf{Xc}_i}_1,
\label{eq:irreg}
\end{equation}
where the operator $\norm{.}_1$ is the \emph{$\ell_1$-norm} and $\mathbf{c}_i$ is the $i$-th column of $\mathbf{C}$.
Our goal is to find the convex combination that best reconstructs each map, in the sense of minimizing the irregularity measure.
Then, we wish to generate map $\mathbf{x}_i$ with a small number of maps.
By using the \emph{$\ell_1$-norm} in \eqref{eq:irreg}, we are accounting for sparse error in the data.

The coefficients are obtained by solving the following optimization problem
\begin{align}
	\min_{\mathbf{C}}\quad & \mathcal{I} \label{eq:mfs} \\
	\textup{s.t.}\quad & diag(\mathbf{C})=0\nonumber \\
	& \mathbf{1}_m^T \mathbf{C} = \mathbf{1}_m^T \nonumber\\
	& \mathbf{C} \geq \mathbf{0}_{m\times m},\nonumber 
\end{align}
where
\begin{equation}
	\mathcal{I}=\norm{\mathbf{X-XC}}_{1} 
\end{equation}
is the global irregularity measure of the data and operator $\norm{.}_{1}$ is the $L_{1}$ matrix norm. The solution is computed in parallel by solving for each $\mathbf{c}_i$ separately \cite{oat2016mfs}.

\begin{figure}[tbh]
\centering
\includegraphics[width=0.43\textwidth]{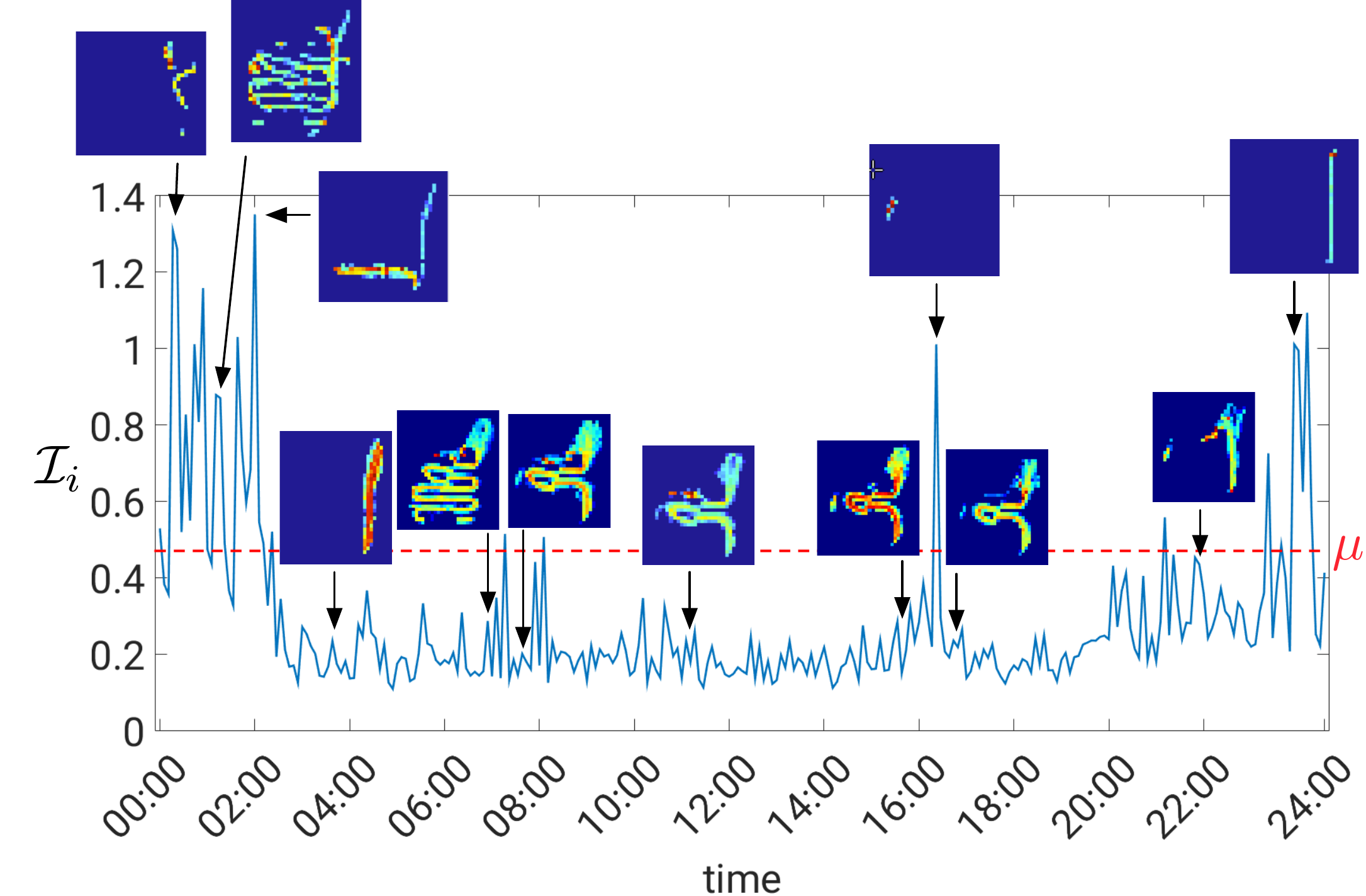}
\caption{Irregularity measure for a complete day. Some of the corresponding occupancy maps are presented to show that higher $\mathcal{I}_i$ values correspond to more peculiar patterns.}
\label{fig:irreg}
\end{figure}

By solving \eqref{eq:mfs} we generate the map $\mathbf{x}_i$ with the other maps in $\mathbf{X}$. 
However, when the error $\mathcal{I}_i$ is too large, a map is an outlier (a unique pattern). 
Therefore, the irregularity $\mathcal{I}_i$ is in fact a measure of uniqueness of the map $\mathbf{x}_i$ within the data set. 
Fig. \ref{fig:irreg} shows the irregularity measure, $\mathcal{I}_i$, for $m$ occupancy maps corresponding to a full day.
As expected, maps with lower $\mathcal{I}_i$ are maps capturing typical states of the queue. 
On the other hand, large values correspond to the occasional patterns.

Choosing a value of $p$, we label as \emph{irregular} the $p\%$\footnote{This is equivalent to apply a threshold $\mu$ to $\mathcal{I}_i$\cite{oat2016mfs}.} of maps with highest irregularity and the others as \emph{regular}\footnote{Note that in \cite{oat2016mfs} these irregular observations correspond to physical keypoints of a shape/model. 
Whereas here we interpret them as maps corresponding to unique patterns in the data. }. With the split and merge approach, presented in next section, the methodology is robust to a large range of $p$ values.

\subsection{Clustering Regular Maps}
\label{sec:clustering}
Solving \eqref{eq:mfs}, we are able to identify regular and irregular (outlier) patterns. Filtering out outliers, the data can be clustered using common clustering methods. Here, we take advantage of the sparse coding framework.
Note that $\mathbf{C} + \mathbf{C}^T$ defines an undirected graph where the nodes represent the maps and edges represent association (non-null coefficients) between maps.
Because, the number of classes $K$ is unknown \emph{a priori}, we estimate $K$ by estimating the number of zero eigenvalues of the Laplacian of this graph.
However, small clusters (with few points) have small impact in the eigenvalues and, when such clusters exist, $K$ is underestimated. 
To retrieve the smaller clusters, we segment the data into a larger number, $\gamma$, of linear subspaces and use a measure intrinsic to the model to compute the distance between the subspaces. If this distance is small, clusters are merged.

The criterion we use is the Normalized Subspace Inclusion (NSI), a distance between subspaces of any dimension \cite{silva2009normalized}.
Formally, it is defined as
\begin{equation}
    NSI(\mathcal{L}_1,\mathcal{L}_2) = \frac{tr(\mathbf{U_1^T U_2 U_2^T U_1})}{\min (d_1,d_2)},
\end{equation}
where $\mathcal{L}_1$ and $\mathcal{L}_2$ are linear subspaces in $\mathbb{R}^n$, $\mathbf{U}_1$ and $\mathbf{U}_2$ are their orthonormal basis and $d_1$, $d_2$ are the subspaces dimensions.
This criterion measures inclusion of subspaces, generalizing the angle between two subspaces.
Then, choosing this criterion comes naturally from the assumptions we made previously, that the data lies in the union of linear subspaces. 
    Similarly, and to exploit this subspace structure in all the clustering process, we over-segment the regular data using \emph{Spectral Clustering} with $\mathbf{C+C^T}$ as adjacency matrix.

\begin{figure}[thb]
\centering
\includegraphics[width=0.295\textwidth]{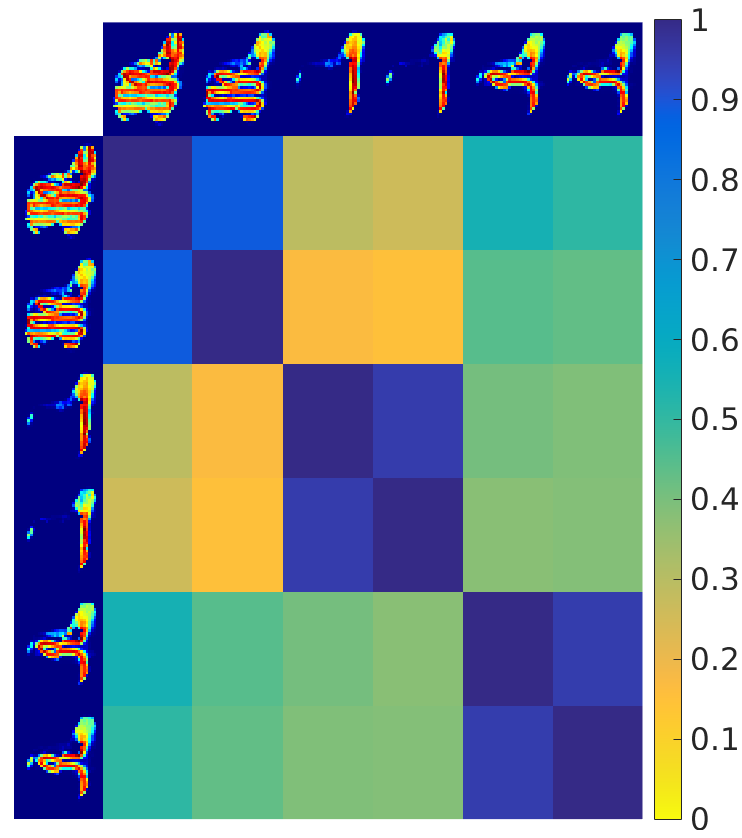}
\caption{NSI affinity matrix for six exemplifying clusters. For each cluster, we show a map belonging to it. Pairs of clusters corresponding to similar configuration have NSI value close to $1$.}
\label{fig:nsi-matrix}
\end{figure}

Fig. \ref{fig:nsi-matrix} shows an affinity matrix with the NSI for six clusters. For each cluster we show here only one of its maps.  
It is clear that the NSI value between clusters that correspond to very similar configurations is close to $1$. 

Clusters are merged if their pair-wise NSI value is above a threshold.
Similarly, after this step, we reassess the maps labeled as \emph{irregular}. We compute the pair-wise NSI affinity for each irregular map and cluster. The map is unified to the cluster with larger affinity if the value is above a threshold, otherwise, the maps remain labeled \emph{irregular}.
This consolidation step, allows a large range for $p\%$ in the selection of irregular/regular patterns. 
\subsection{Methodology Summary}

The methodology proposed in this section is summarized in Algorithm \ref{alg:method}.
First, we solve \eqref{eq:mfs} to compute the coefficients that relate the patterns (line 1) and then we rank the maps by irregularity to identify regular and irregular maps (line 2). Next, we estimate the number of clusters and partition regular data accordingly (lines 3 and 4). Then, clusters are consolidated using the affinity measure (lines 5 and 6). Finally, irregular maps are classified according to the consolidated clusters (line 7). The output is the labeling for all data points and the corresponding partition into regular and irregular patterns.

Next, we present results of applying this methodology to real data.
\begin{algorithm}
\caption{Identifying and clustering regular occupancy maps in the data.}
\label{alg:method}
\begin{algorithmic}[1]
\Statex \textbf{Input:} $\mathbf{X}$ - Data matrix
\Statex \qquad\quad $p$ - \% of maps to label as irregular
\Statex \qquad\quad $th$ - NSI threshold
\Statex \textbf{Output:} $\mathbf{X}_{reg}$ - Regular data
\Statex \qquad\quad\quad\!$\mathbf{X}_{irreg}$ - Irregular data
\Statex \qquad\quad\quad\!\!$cl_{reg}$ - Regular data class labels
\Statex
\Statex \emph{Step 1: Identifying Regular and Irregular Maps}
\State $\mathcal{I} \gets \text{Solve \eqref{eq:mfs} for } \mathbf{C}$
\State $\mathbf{X}_{reg}, \mathbf{X}_{irreg} \gets $ Divide regular and irregular data 
\Statex \hspace{23.3mm} according to $p$ and $\mathcal{I}$
\Statex
\Statex \emph{Step 2: Clustering Regular Maps}
\State $\gamma \gets$ Estimate number of clusters
\State $cl_{reg} \gets $ Segment $\mathbf{X}_{reg}$ into $\gamma$ clusters \label{lin:segm}
\State $affinity \gets $ Compute NSI between $\gamma$ clusters
\State $cl_{NSI} \gets $ Merge clusters in $cl_{reg}$ with affinity $> th_{NSI}$
\State $cl_{reg},\mathbf{X}_{reg},\mathbf{X}_{irreg}\gets $ Reclassify $\mathbf{X}_{irreg}$ according 
\Statex \hspace{33mm}to $cl_{NSI}$
\end{algorithmic}
\end{algorithm}

\section{Experimental Results}
\label{sec:results}
In this section, we present quantitative results for synthetic data and qualitative results for real data, with periods of one day and fourteen days. %and four days during Christmas time. 
We used seven depth cameras to cover the X-ray queue area at the Lisbon international airport. The descriptor used integrates $5min$ occupancy.
\subsection{Synthetic Data}
To quantitatively assess the performance of our method and to compare it with other clustering methods, we perform a set of experiments with synthetic data. 
The classes are unknown to the algorithm and the data is corrupted with noise and outliers.

\begin{figure}[bht]
\centering
\subfloat[Regular maps]{\includegraphics[width=0.45\textwidth]{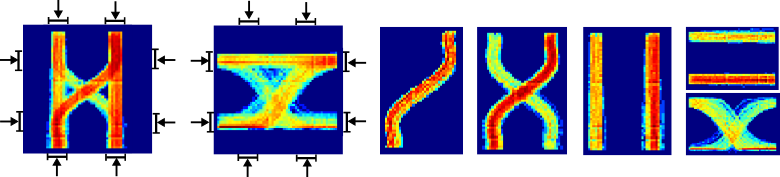}
\label{fig:synth_maps}}\\
\hspace{0.3mm}
\subfloat[Outliers]{\includegraphics[width=0.44\textwidth]{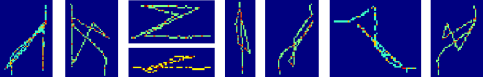}
\label{fig:synth_out}}
\caption{Examples of occupancy maps generated from synthetic data: (a) regular maps, with the 8 doors illustrated in the two left maps; (b) outlier maps. Each map integrates data from a period of $5min$.}
\label{fig:synth}
\end{figure}

We consider a squared space of $400m^2$, with a total of 8 doors and several possible paths between those doors (Fig. \ref{fig:synth_maps}). At each time instant, we randomly generate a simulated passenger at a given door. Each passenger moves along a set of predefined paths (subject to perturbations), with speed $s\backsim\mathcal{N}(1.4ms^{-1},0.05)$. Similar to the method for people detection used in the real setup \cite{carvalho2016detecting}, objects are correctly detected $96\%$ of time. 
Outlier trajectories are created with passengers partially following one of the regular paths but then temporarily moving away to go through random points in space, returning finally to the normal course. 
We build $5min$ occupancy maps as explained in Section \ref{sec:occ_map}, with one binary map per second.
Fig. \ref{fig:synth} shows some examples of regular and irregular occupancy maps created with this synthetic data. 
These data recreates the real setting, with a similar large space with several doors and paths, passengers moving at the average human speed and occupancy maps with the same characteristics.
In the following experiments, we create data matrices with 10 clusters, corrupted with several percentages of outliers. For each such percentage, we run 20 simulated experiments with 576 maps each. 

We compare our method with three well established clustering approaches: \emph{K-Means} \cite{lloyd1982least}, \emph{K-Medoids} \cite{kaufman1987clustering} and DBSCAN \cite{ester1996density}. 
We include K-Means because it is widely used, however, it is inadequate for this application, performing poorly.
We evaluate the clustering error, defined as the rate between incorrectly classified points and the total number of points. In this comparison, we consider only the error in the regular maps, meaning we do not assess the explicit identification of irregular maps. 
We input the correct number of clusters to K-Means and K-Medoids, and use angle between points as distance function.
Table \ref{tab:ec} shows the average clustering error as a function of the percentage of outliers. DBSCAN accounts for outliers in the data but, similar to K-Means, performs poorly with this data. K-Medoids achieves small errors for lower percentages but loses accuracy as the number of outliers increases.
On the other hand, our approach is able to achieve error close to zero for all percentages of outliers. Note that our error is higher with $20\%$ of outliers because we set $p=0.50$ and, for some experiments, that value of $p$ removes classes from the regular data. With a smaller $p$, we can also achieve zero error for this amount of outliers.
\begin{table}[hbt]
\caption{Average clustering error as a function of the percentage of outliers, $p_{out}$. For each trial with K-Means and K-Medoids, we used the best of 10 replicates. For our method, we report the error for $p=0.50$, NSI threshold $0.93$ and $\gamma = 17$ clusters.}
\label{tab:ec}
\centering
\begin{tabular}{llllllllll}
\toprule
$p_{out}$& 0.20&0.30&0.40&0.50&0.60\\
\midrule
K-Means &0.284&0.334&0.388&0.393&0.402\\
DBSCAN&0.206&0.223&0.228&0.223&0.212\\
K-Medoids&\textbf{0}&0.030&0.030&0.067&0.104\\
Proposed Method  &0.022&\textbf{0.009}&\textbf{0.004}&\textbf{0.003}&\textbf{0.005}\\
\bottomrule
\end{tabular}
%}
\end{table}

\begin{figure}[bht]
\centering
\includegraphics[width=0.48\textwidth]{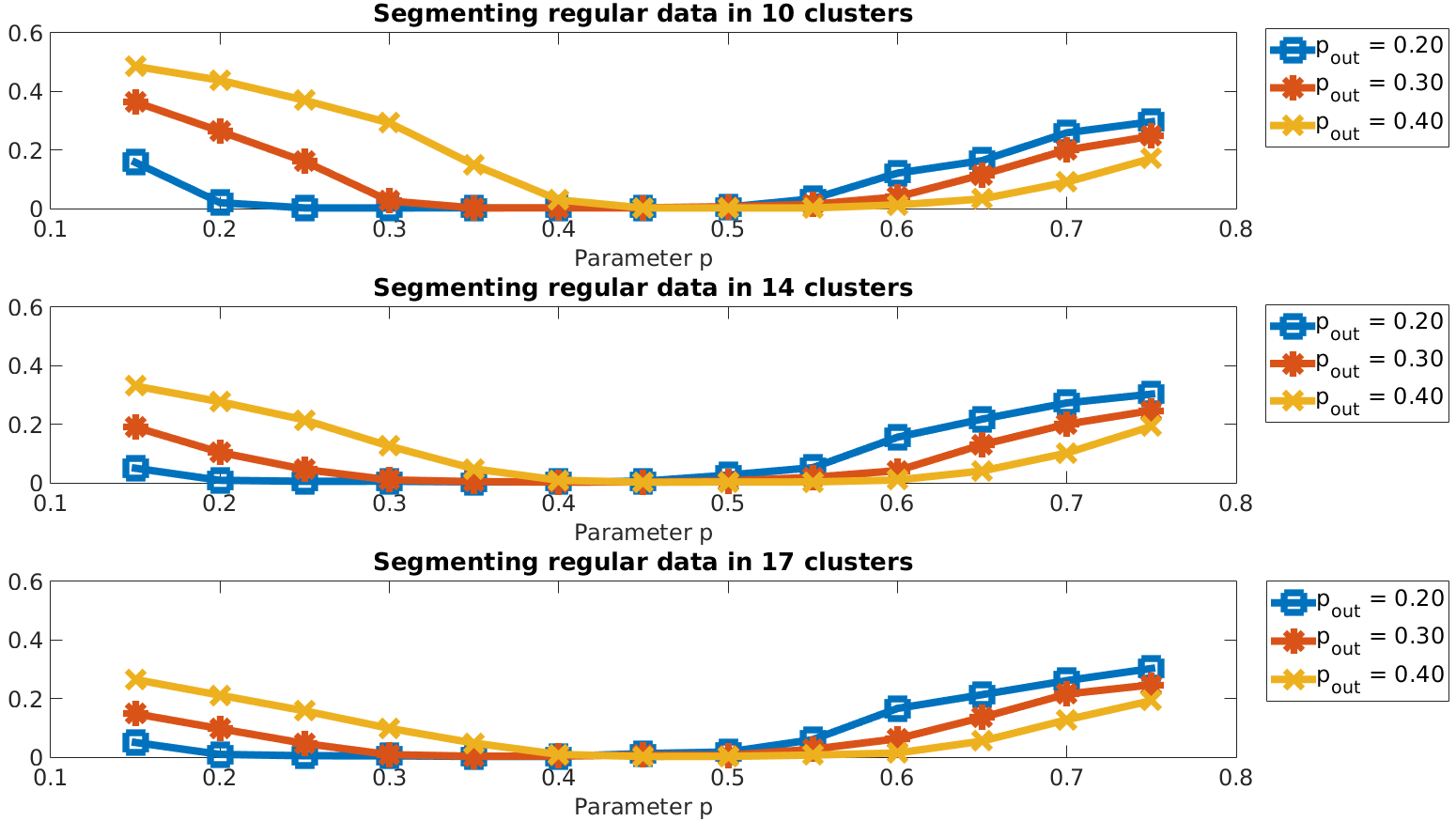}
\caption{Clustering error versus parameter $p$ for three levels of outliers. Top plot: segmenting data in 10 clusters. 
Middle plot: segmenting in 14 clusters. Bottom plot: segmenting in 17 clusters. In each plot we show the error for three percentages of outliers, $p_{out} \in \{0.20, 0.30, 0.40\}$.}
\label{fig:ec_p_splits}
\end{figure}

In Fig. \ref{fig:ec_p_splits}, we evaluate the clustering error as a function of the parameter $p$, percentage of outliers and as function of number of clusters in which the regular maps are segmented, $\gamma$. 
As we show, our method has the best performance for a large range of values of parameter $p$. 
To achieve this, we must ensure that all outliers are removed, $p > \%outliers$, and that none of the clusters is removed from the regular data.
On the other hand, even if not all outliers are removed, the split and merge strategy improves the results because the remaining outliers have a reduced impact on the original clusters. 
\subsection{Real Data: Analyzing One Day}
\label{sec:one-day}
\begin{figure}[ht]
\centering
\subfloat[Coefficients matrix $\mathbf{C}$.]{\includegraphics[width=0.33\textwidth]{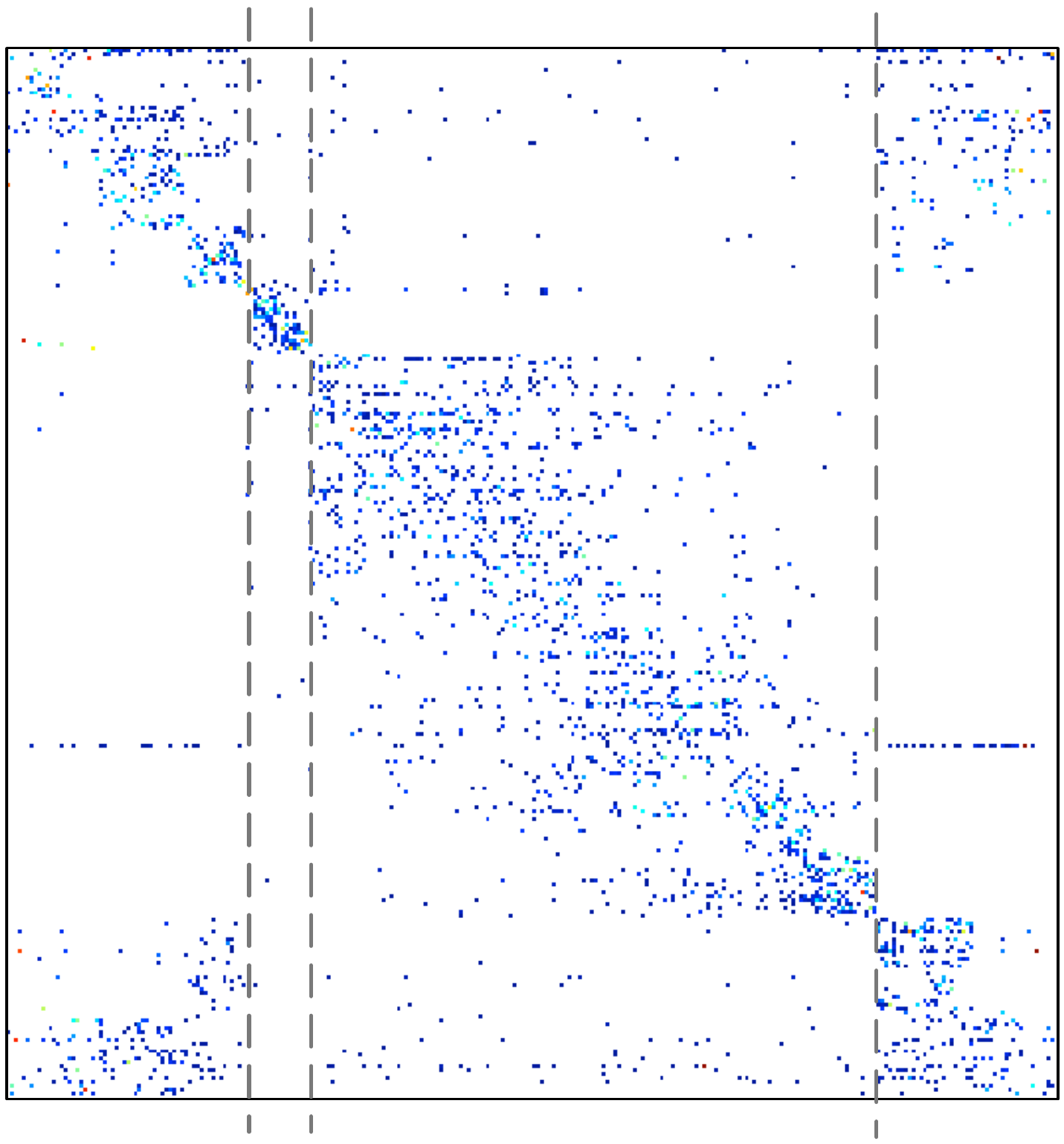}
\label{fig:c-16-03}}\\
\subfloat[Clusters from the Spectral Clustering with $\gamma=5$.]{\includegraphics[width=0.33\textwidth]{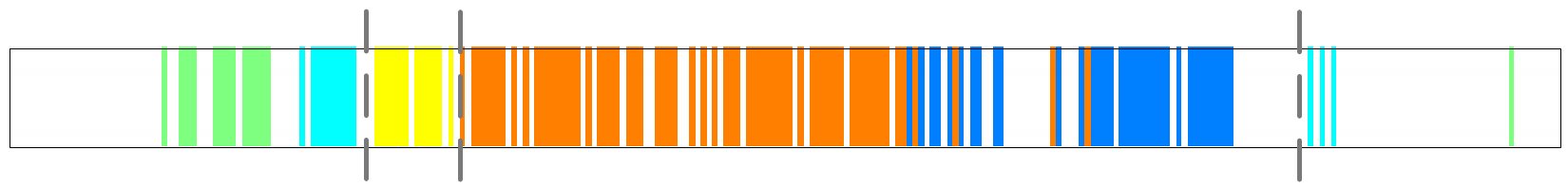}
\label{fig:spc-16-03}}\\
\subfloat[Clusters after merging with NSI.]{\includegraphics[width=0.33\textwidth]{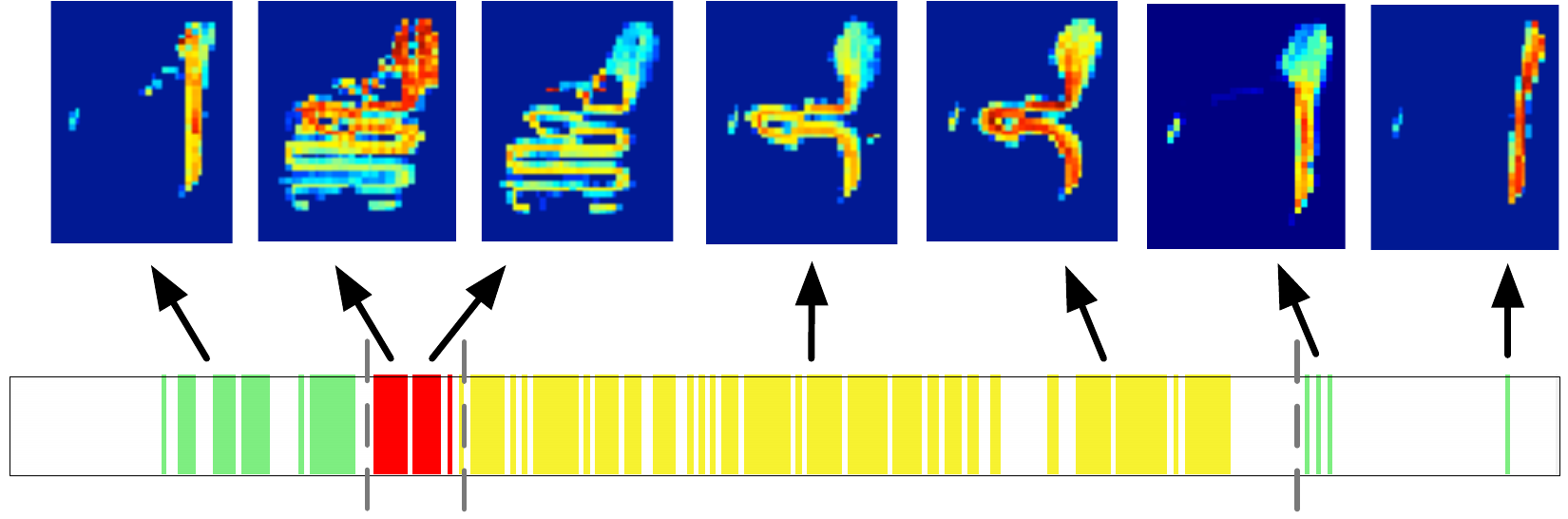}
\label{fig:nsic-16-03}}\\
\hspace{-0.5mm}
\subfloat[Final clusters.]{\includegraphics[width=0.35\textwidth]{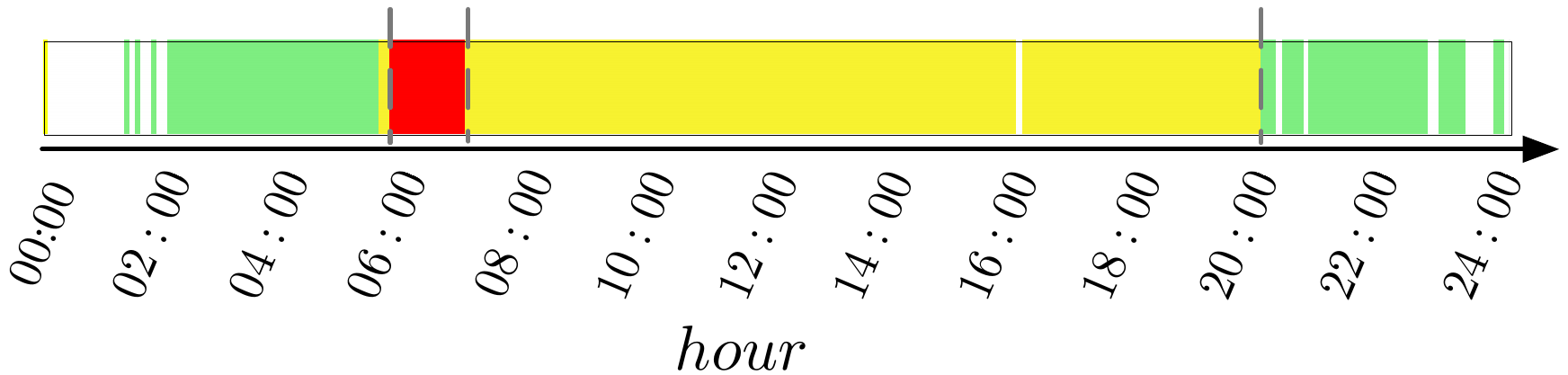}
\label{fig:final-16-03}}
\caption{Results for one day of data: (a) coefficients matrix $\mathbf{C}$; (b) over-segmentation after labeling $50\%$ of the maps as irregular; result for $\gamma=5$, with one color per cluster (white for irregular maps); (c) consolidated clusters obtained using the NSI criterion, with one color per cluster, and some of the corresponding regular maps; (c) final clusters, obtained after classifying maps initially labeled as irregular. The gray vertical dashed line in all sub-figures allows to relate coefficients in $\mathbf{C}$ with clustering results.}
\label{fig:16-03}
\end{figure}

Here we present in detail the results of each step of our methodology applied to real data of one day. Fig. \ref{fig:c-16-03} shows the $\mathbf{C}$ matrix obtained by solving \eqref{eq:mfs} for this day.
The block diagonal shape of $\mathbf{C}$ suggests that classes appear in sequence and do not change often during the day.

Fig. \ref{fig:irreg} shows the \emph{irregularity} for this sequence of maps, given this matrix $\mathbf{C}$.
After ranking the maps, we remove the $50\%$ with highest irregularity.
Next, we partition the \emph{regular} data ($50\%$ with lowest error) in $5$ clusters using spectral clustering\footnote{The adjacency matrix for the Spectral Clustering is $\mathbf{C}+\mathbf{C}^T$.}. 
Fig. \ref{fig:spc-16-03} shows the obtained labels, with one color per cluster and irregular maps in white.
By consolidating these clusters with the NSI criterion, we obtained three clusters (green, yellow and red labels) corresponding to three different queue configurations, depicted in Fig. \ref{fig:nsic-16-03}.
Finally, we classified the $50\%$ maps previously labeled as irregular to assess if they belong to any regular clusters.
Fig. \ref{fig:final-16-03} shows the final labels, where $11\%$ of the maps remain labeled \emph{irregular}.

During the periods 00:00$am$ - 06:15$am$ and 07:45$pm$ - 11:59$pm$, the queue had low inflow of passengers and people would go straight from the entrance to the exit of the queue (green label in Fig.\! \ref{fig:final-16-03}). 
The period with largest passenger inflow occurs between 06:15$am$ and 07:15$am$ (red label). In this period, the queue structure is more complex, occupying a larger area to accommodate all passengers. 
Yellow label corresponds to a configuration with moderate flow.
Irregular maps, with white label, occur mainly in the night periods (with low passenger inflow), when cleaning and maintenance operations are performed.

\subsection{Real Data: Analyzing an Extended Time Period}
\label{sec:larger-periods}
\begin{figure*}[tbh]
\centering
\includegraphics[width=0.8\textwidth]{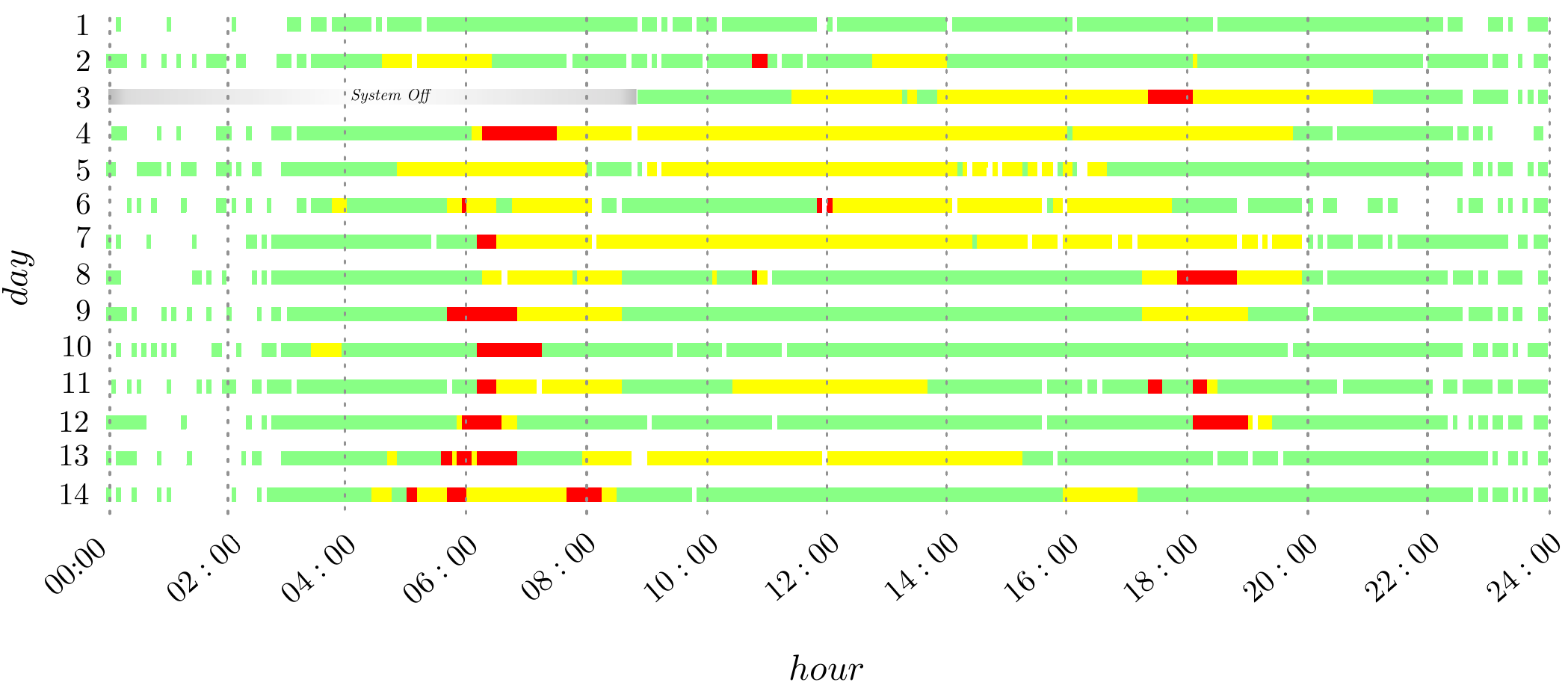}
\caption{Results of our methodology for 14 days of data, with $\mathbf{X} \in \mathbb{R}^{d \times 3583}$. Days are shown separately for easier analysis. Each queue configuration is labeled with a different color, with white corresponding to irregular periods. Day 3 has a large period without labeling, identified in gray, because the acquisition system was off.}
\label{fig:nsic-14days}
\end{figure*}
Fig. \ref{fig:nsic-14days} shows results for $14$ days of operation. The color labeling is shown for each day separately, although we applied the methodology to all data, with $\mathbf{X} \in \mathbb{R}^{d \times 3583}$.
These colors correspond to the same configurations of Fig. \ref{fig:16-03}.
White label is associated with irregular or empty maps\footnote{Empty maps were not included in the data matrix.}.

\begin{figure}[bht]
\centering
\includegraphics[width=0.38\textwidth]{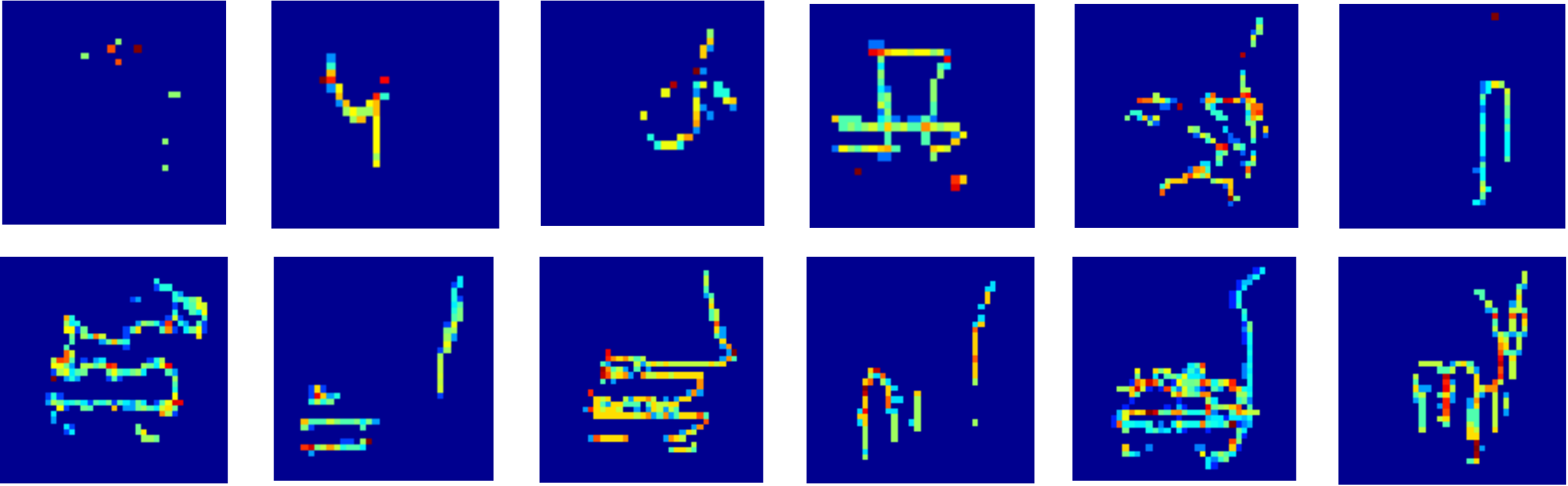}
\caption{Some of the irregular maps identified within a 14 day period.}
\label{fig:irreg-14days}
%\label{fig:pcl-rx}
\end{figure}

Although all days are different from each other, periods with more inflow look similar: between 06:00$am$ and 07:00$am$ or between 06:00$pm$ and 07:00$pm$. 
Irregular periods of time, as the ones depicted in Fig. \ref{fig:irreg-14days}, occur mainly in the early or late hours of the day and represent $12\%$ of the fourteen days, a time lapse of around $39$ hours\footnote{Day $3$ has a large period of time without labeling because the acquisition system was off.}.
Without identifying the outliers, the clusters computed by common approaches are contaminated and the corresponding subspaces may not represent the meaningful classes.

\begin{figure}[bht]
\centering
\includegraphics[width=0.4\textwidth]{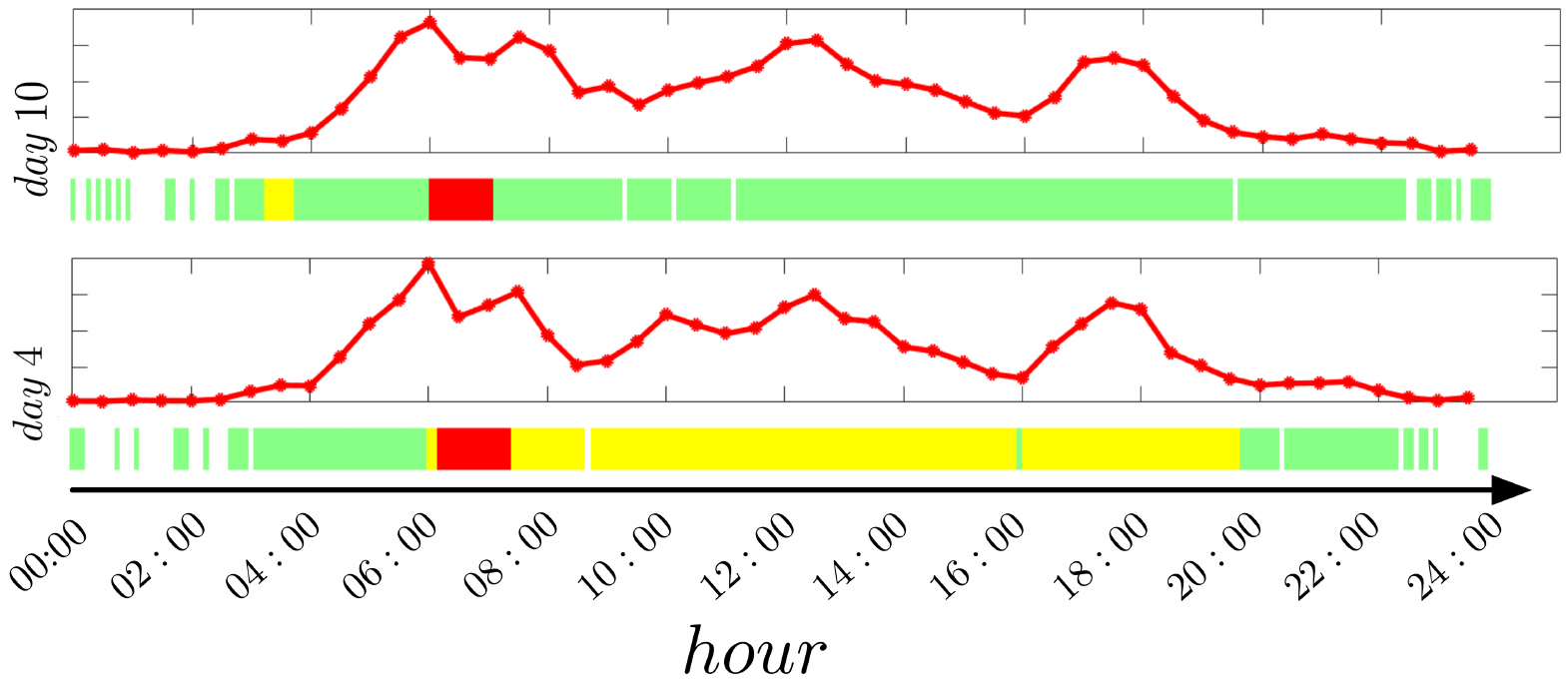}
\caption{Counting and labeling for two different days. Although the counting is similar, the days are different.}
\label{fig:counts-vs-labels}
\end{figure}

As we previously showed in Fig. \ref{fig:countings-maps}, similar passenger count at specific periods corresponds to different queue patterns. Fig. \ref{fig:counts-vs-labels} shows the same occurs for full days.
This is due to several factors affecting the state of the queue, such as the number of open X-ray gates, number of staff members operating or luggage contents, for example. 
In other words, for the same number of passengers, the service can be operated differently.

\section{Discussion}
\label{sec:disc}
We propose an unsupervised approach with few parameters to setup.
The first parameter is the period of time covered by each occupancy map, \emph{e.g.}, $30s, 1min, 5min$. 
This period of time should be chosen depending on the activity we want to capture with the descriptor.
For these experiments, we empirically chose $5min$ maps, that showed to be a good trade-off between computational cost and data filtering.
Another parameter is the period of time in which we search for clusters. 
In this paper, we use one day and two weeks but we can choose shorter or longer periods.
Our methodology presents consistent results and similar clusters can be identified whether analyzing a day by itself or combined with other days, including smaller clusters. 

\begin{figure}[bht]
\centering
\includegraphics[width=0.38\textwidth]{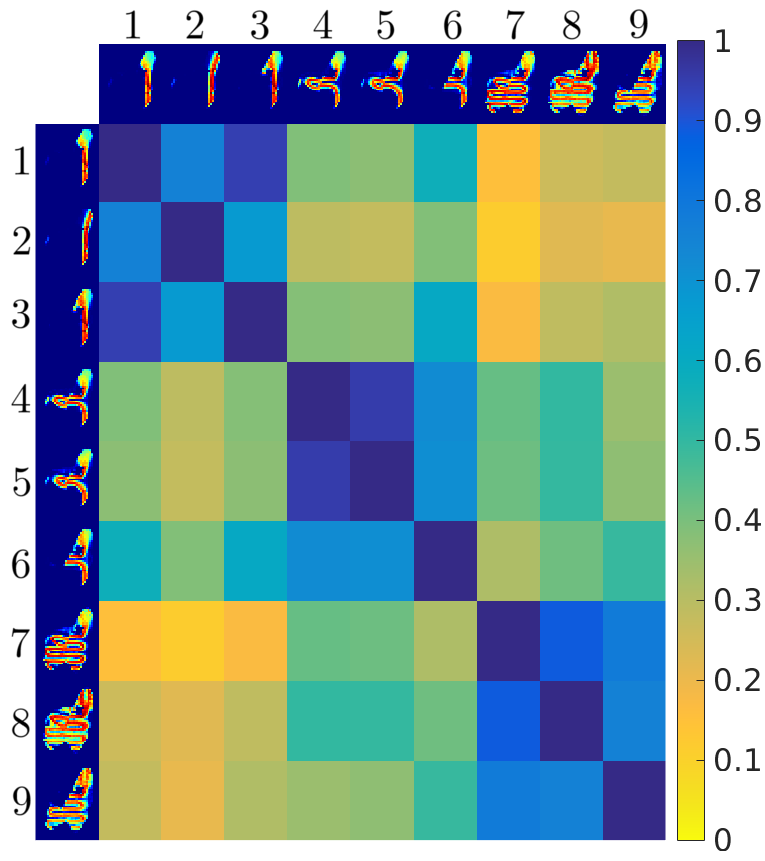}
\caption{NSI affinity matrix including maps with missing data. The characteristics of the descriptor and NSI allow for a smooth degradation of the affinity value when missing data.}
\label{fig:affinity-missing}
\end{figure}

Fig. \ref{fig:affinity-missing} shows the affinity matrix for nine clusters in a period where one of the (seven) cameras broke. For each cluster, we show the mean of its maps.
Clusters $7$, $8$ and $9$ are similar except in the area where the camera failed, in the middle left side of $9$. Similarly, cluster $6$ corresponds to the same queue state as clusters $4$ and $5$. 
Our method degrades gracefully in the presence of perturbations and clusters $6$ and $9$ have high affinity with the corresponding configuration classes.

Although the methodology presented in this paper is unsupervised, when working with real data and operational scenarios, interaction with end users/experts is very useful.
Because of the large amount of data, airport experts are not able to describe classes \emph{a priori}.
By providing only representative patterns, our methodology works as a filter and saves users from the overwhelming load of interpreting all data.
When presented with the relevant patterns, the airport experts recognize the classes as meaningful and understand the affinity matrix.
Finally, with these classes, and applying queueing and discrete events theory, we are able to compute individual and more accurate models for each state.
\section{Conclusion}
\label{sec:conclusion}
We proposed a new descriptor to characterize the flow of people in large public infrastructures.
Based on 3D data and without tracking people, we provided an occupancy map of the space---a very useful indicator of the service performance.
We proposed an unsupervised methodology to identify and cluster these occupancy maps, without knowing their number or shape/configuration \emph{a priori}.
The approach is divided into two main steps: identify regular and irregular periods/maps, and cluster the regular maps into classes corresponding to the different queue configurations.
This methodology proposes a continuous irregularity measure for each map. The first step of the approach uses this measure as a means to identify the most abnormal maps in the data set. 
The second step segments the regular maps, based on an estimate of the number of clusters, and then consolidates the clusters using the Normalized Subspace Inclusion (NSI) criterion.

The approach gave consistent results, independently of the quantity of data analyzed, with the descriptor and NSI having smooth deterioration in the presence of missing data.

The data originates from different days but it has a common pattern: the busiest hours occur in the same periods of the day.
Finally, irregular maps exist mainly during late night, when maintenance operations are due.

As future work, we plan to augment the model to account for more information sources, \emph{e.g.}, velocity, flight schedule, flight destiny, in order to strengthen the assessment of the service performance.
\section*{Acknowledgments}
\label{sec:ack}
This work was done in collaboration with Thales Portugal S.A. and ANA \textit{Aeroportos de Portugal}, within the SMART-er project.
The authors would like to specially thank Jo\~ao Mira (Thales), Francisco Oliveira and Isabel Rebelo (ANA).
%

%\bibliography{References}
\bibliography{paper-t-its2017}
\bibliographystyle{/usr/share/texmf-dist/bibtex/bst/IEEEtran/IEEEtran}
 
\begin{IEEEbiography}[{\includegraphics[width=1in,height=1.25in,clip,keepaspectratio]{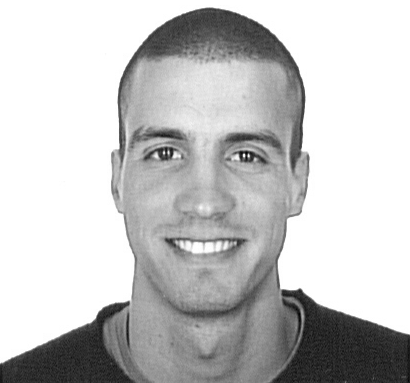}}]{João Carvalho}
   holds a M.Sc.\ degree (2012) in Electrical and Computer Engineering from Instituto Superior Técnico (IST), Lisbon, Portugal. He is a Ph.D.\ student at IST under the Networked Cyber Physical Systems (NetSyS) doctoral program. Since 2011, he is a Research Assistant at the Institute for Systems and Robotics, Lisbon, Portugal. Between 2013 and 2014, he was a Research Assistant at Thales Portugal. His research interests include computer vision and robotics.
 \end{IEEEbiography}
 
\begin{IEEEbiography}[{\includegraphics[width=1in,height=1.25in,clip,keepaspectratio]{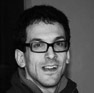}}]{Manuel Marques}
    received a Portuguese Science Foundation (FCT) Fellowship and is a researcher at Instituto de Sistemas e Robótica (ISR),  Instituto Superior Técnico (IST). He holds a Ph.D.\ in Electrical and Computer Eng from IST (2011) and is his main areas of research include 3D reconstruction, object recognition and video processing. Recently he focus on applications to transportation and city mobility. He his principal investigator of several projects including automatic survey of risk for cycling in cities using computer vision. 
 \end{IEEEbiography}
 
\begin{IEEEbiography}[{\includegraphics[width=1in,height=1.25in,clip,keepaspectratio]{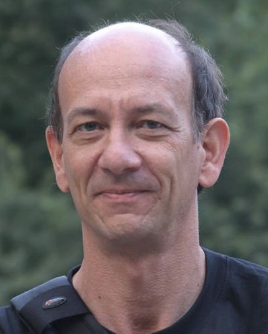}}]{João Paulo Costeira}
    is an Associate Professor in the Electrical and Computer Engineering Department at Instituto Superior Técnico (IST). He is also a researcher at the Instituto de Sistemas e Robótica (ISR) where he coordinates the Signal and Image Processing Group. His research interests are focused in the area of computer vision with special interests on 3D reconstruction from video, image matching and recognition. He holds a Ph.D.\ in Electrical and Computer Eng from IST (1995) and he was a visiting scientist at the Vision and Autonomous Systems Center(VASC), Carnegie Mellon University from 1992-1995. Currently he is scientific director of the CMU-Portugal program, a joint venture between the Portuguese Government and Carnegie Mellon University. %Also he coordinates the dual PhD program in Electrical and Computer Eng between IST and CMU.
 \end{IEEEbiography}

\vfill
\end{document}